\pdfoutput=1

\documentclass[11pt]{article}

\usepackage[final]{acl}

\usepackage{times}
\usepackage{latexsym}
\usepackage{acl}
\usepackage{amsmath}
\usepackage[T1]{fontenc}

\usepackage[utf8]{inputenc}

\usepackage{microtype}

\usepackage{inconsolata}

\usepackage{graphicx}
\usepackage{multirow}

\usepackage{graphics}
\usepackage{graphicx}
\usepackage{xspace}
\usepackage{subcaption}
\usepackage{amsmath}
\usepackage{amsthm}
\usepackage{mathrsfs}
\usepackage{array}
\usepackage{textcomp}
\usepackage{weiwAlgorithm}
\usepackage{url}
\usepackage{diagbox}
\usepackage{color}
\usepackage{booktabs}
\usepackage{listings}
\usepackage{tcolorbox}
\usepackage{setspace}
\usepackage{multirow}
\usepackage{booktabs}    
\usepackage{tabularx}    
\usepackage{array}       
\usepackage{makecell}    
\usepackage{enumitem}
\usepackage{lineno}
\usepackage{float}    
\usepackage{placeins} 
\usepackage{stfloats} 
\usepackage{tabularx}
\usepackage{array}
\usepackage{tabularx}
\usepackage{array}
\usepackage{booktabs}   
\usepackage{xcolor}     
\usepackage{amsmath}    
\usepackage{amssymb}    
\usepackage{tabularx}
\usepackage{array}
\usepackage{booktabs}
\usepackage{subcaption}   
\usepackage{xcolor}
\usepackage{amsmath}

\definecolor{kgblue}{RGB}{55, 115, 179}
\definecolor{wikigreen}{RGB}{46, 153, 83}
\definecolor{weborange}{RGB}{220, 130, 30}

\usepackage[normalem]{ulem}
\useunder{\uline}{\ul}{}

\newtheorem{definition}{Definition}

\newcommand{\srag}{HydraRAG\xspace}

\newcommand{\myparagraph}[1]{\noindent \textbf{#1}.}

\newcommand{\myparagraphunderline}[1]{\noindent \underline{#1.}}
\newcommand{\myparagraphquestion}[1]{\noindent \textbf{#1?}}
\newcommand{\myparagraphunderlinenew}[1]{\noindent \underline{#1,}}
\newcommand{\RomanNumeralCaps}[1]
    {\MakeUppercase{\romannumeral #1}}
\newcommand{\ie}{{i.e.,}\xspace}

\usepackage{graphicx}
\providecommand{\texorpdfstring}[2]{#1}     

\newcommand{\titlelogo}{%
  \raisebox{-0.2\height}{\includegraphics[height=1.6em]{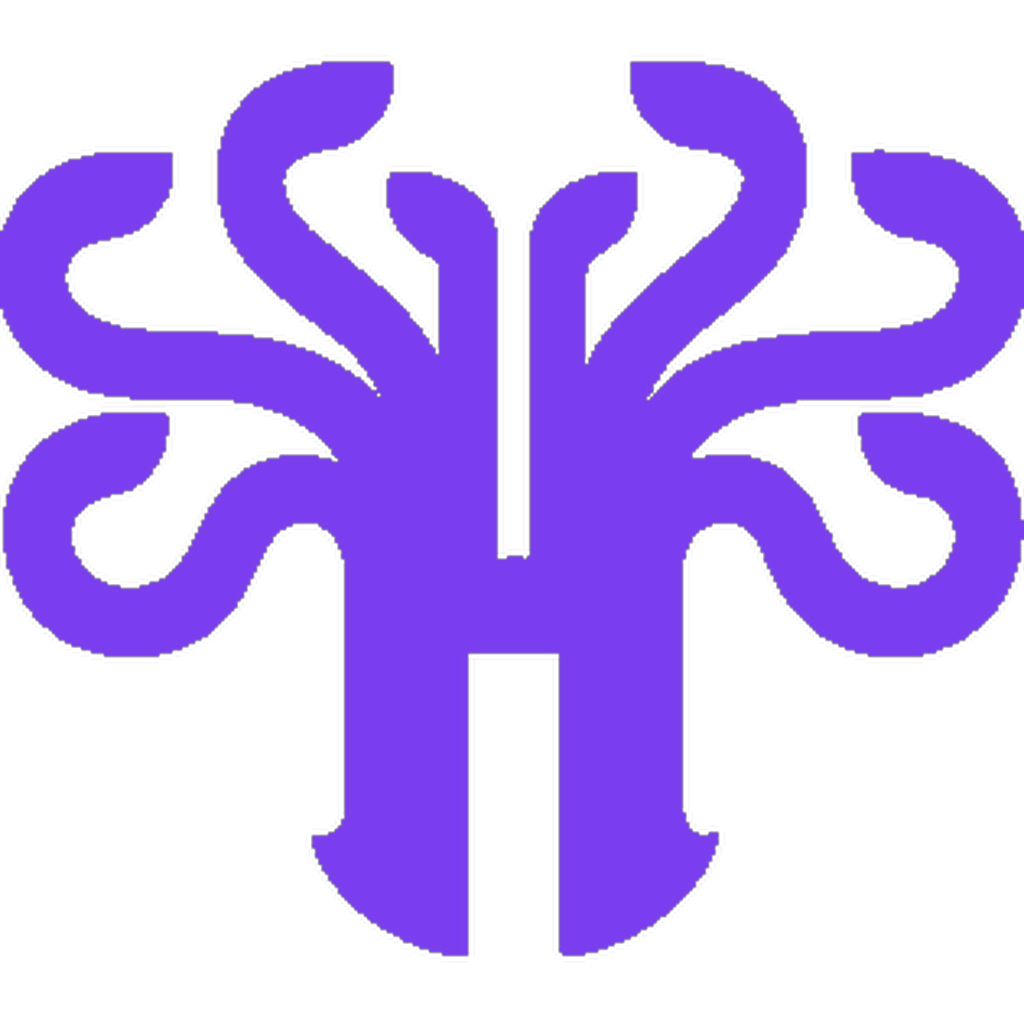}}%
}

\title{\texorpdfstring{%
  \titlelogo\ \srag: Structured Cross-Source Enhanced\\  Large Language Model Reasoning%
}{Hydra: Structured Cross-Source Enhanced Large Language Model Reasoning}}

\usepackage{graphicx}
\usepackage{titling}

\author{
 \textbf{Xingyu Tan\textsuperscript{1,2}},
 \textbf{Xiaoyang Wang\textsuperscript{1,*}},
 \textbf{Qing Liu\textsuperscript{2}},
 \textbf{Xiwei Xu\textsuperscript{2}},
\\
 \textbf{Xin Yuan\textsuperscript{2}}, 
 \textbf{Liming Zhu\textsuperscript{2}},
 \textbf{Wenjie Zhang\textsuperscript{1}}
\\
 \textsuperscript{1}University of New South Wales, Australia\\
 \textsuperscript{2}Data61, CSIRO, Australia
\\
   \texttt{\{xingyu.tan, xiaoyang.wang1, wenjie.zhang\}@unsw.edu.au}\\
   \texttt{\{q.liu, xiwei.xu, xin.yuan, liming.zhu\}@data61.csiro.au}
}

\begin{document}
\maketitle
\begingroup
\def\thefootnote{$*$}
\footnotetext{\raggedright Corresponding author.}
\endgroup

\begin{abstract}

Retrieval-augmented generation (RAG) enhances large language models (LLMs) by incorporating external knowledge. Current hybrid RAG system retrieves evidence from both knowledge graphs (KGs) and text documents to support LLM reasoning. However, it faces challenges like handling multi-hop reasoning, multi-entity questions, multi-source verification, and effective graph utilization. To address these limitations, we present \textbf{HydraRAG}, a training-free framework that unifies graph topology, document semantics, and source reliability to support deep, faithful reasoning in LLMs. HydraRAG handles multi-hop and multi-entity problems through agent-driven exploration that combines structured and unstructured retrieval, increasing both diversity and precision of evidence. To tackle multi-source verification, HydraRAG uses a tri-factor cross-source verification (source trustworthiness assessment, cross-source corroboration, and entity-path alignment), to balance topic relevance with cross-modal agreement. By leveraging graph structure, HydraRAG fuses heterogeneous sources, guides efficient exploration, and prunes noise early. Comprehensive experiments on seven benchmark datasets show that HydraRAG achieves overall state-of-the-art results on all benchmarks with GPT-3.5, outperforming the strong hybrid baseline ToG-2 by an average of 20.3\% and up to 30.1\%. Furthermore, HydraRAG enables smaller models (e.g., Llama-3.1-8B) to achieve reasoning performance comparable to that of GPT-4-Turbo. The source code is available on \url{https://stevetantan.github.io/HydraRAG/}.

\end{abstract}

\section{Introduction}\label{sec:intro}


\begin{figure*}[t]
    \centering
    \includegraphics[width=0.99 \linewidth]{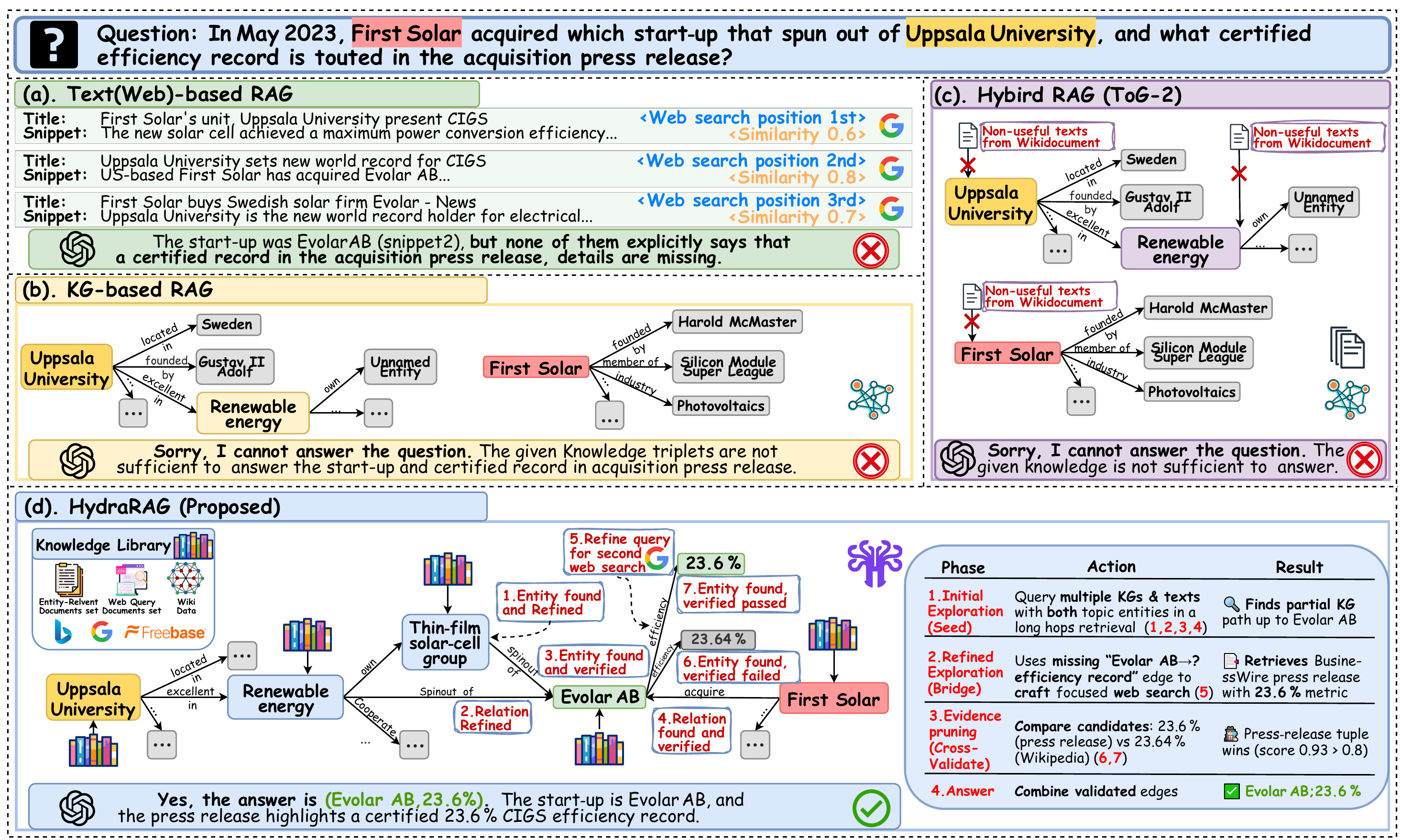}
    \caption{Representative workflow of four LLM reasoning paradigms.}
    \vspace{-2mm}
    \label{fig:intro_demo}
\end{figure*}

Large Language Models (LLMs) have achieved remarkable performance by scaling to billions of parameters and pre‑training on vast and diverse corpora
\cite{brown2020language,chowdhery2023palm}.
However, the prohibitive expense of full-model training for LLMs makes continual retraining infeasible, causing static parametric knowledge to quickly become obsolete and resulting in factual gaps and hallucinations \cite{besta2024graphGoT,touvron2023llama}.
This issue is alleviated by retrieval-augmented generation (RAG), which fetches external evidence at inference time.
\cite{gao2023retrieval}.


Many RAG systems rely on vector retrieval over text, embedding question and documents into a dense space and selecting semantically similar passages \cite{baek2023knowledge,jiang2023structgpt,huang2024embedding,huang2024foundation}. 
While effective for measuring text similarity, such approaches struggle with complex reasoning that requires integrating heterogeneous clues across multiple documents \cite{tog2.0ma2024think}. 
Specifically, (i) different passages may reference distinct entities that share the same underlying concept, such as, Evolar and  Evolar AB in Figure~\ref{fig:intro_demo}(a) refer to the same start‑up company; (ii) a single passage often covers only one facet of an entity, omitting other critical attributes found in other texts or documents.  In Figure~\ref{fig:intro_demo}(a), with the real-time web information implementation, the naive RAG could find the answer to the first part of the question, but could not relate this entity to other text corpora.


To address these challenges, incorporating external knowledge sources, like Knowledge Graphs (KGs), is promising as KGs offer abundant factual knowledge in a structured format, serving as a reliable source to improve LLM capabilities~\cite{tog1.0sun2023think,pogtan2025paths}.
KG-based RAG approaches prompt LLMs with retrieved KG triples or paths relevant to the question, and their effectiveness in dealing with complex reasoning tasks has been demonstrated by researchers  \cite{pogtan2025paths}.
Although they benefit from the structural and factual nature of KGs, they inherently suffer from inner incompleteness, lack of information beyond their ontology, and high cost of updating \cite{tog2.0ma2024think}. 
For example, as shown in Figure~\ref{fig:intro_demo}(b), the KG search is limited by being unable to provide further information about ``Renewable Energy'' and ``First Solar''.
Some recent works focus on integrating text and KG as a hybrid RAG system \cite{li2023chaincok,tog2.0ma2024think}.

\myparagraph{Limitations of existing methods}
Current approaches typically follow a simple \texttt{retrieve-} \texttt{and-select} routine. For instance, CoK alternates between different Knowledge Bases (KBs), choosing one source at each step and retrieving an answer directly \cite{li2023chaincok}. ToG-2, shown in Figure \ref{fig:intro_demo}(c), simultaneously queries text and KG, extracting one-hop triples for each question keyword and using an LLM to select the best answer \cite{tog2.0ma2024think}. This strategy suffers from four limitations:

\myparagraphunderline{Multi-source verification}  
When faced with multiple sources, many approaches simply concatenate evidence and let the LLM decide. This over-relies on the LLM’s semantics without accounting for source reliability or cross-source consistency, leading to both under- and over-pruning of evidence.

\myparagraphunderline{Multi-hop reasoning}  
Existing methods typically retrieve only one-hop relations in text and KG per step and rely on LLMs for semantically relevant candidates pruning. This greedy, local strategy may prune the correct multi-hop path prematurely and fail to consider the global reasoning structure.


\myparagraphunderline{Multi-entity questions} 
Typical pipelines explore each topic entity independently. For questions involving several entities, this produces large candidate sets containing paths unrelated to the other entities, reducing precision and introducing noise.



\myparagraphunderline{Graph structure utilization}
Current methods fetch triples from each source and pass them to the LLM without merging them into a single graph. Lacking this global structure, the LLM cannot perform efficient graph-based exploration or pruning, so all direct neighbors from KGs and text remain, adding substantial noise.

\myparagraph{Contributions}
We present \textbf{\srag}, shown in Figure~\ref{fig:intro_demo}(d), a structured source-aware retrieval-augmented framework that brings together graph topology, document semantics, and source reliability signals to support deep, faithful reasoning in LLMs. Unlike methods that treat KG triples and text passages as separate evidence, \srag extracts joint KG–text reasoning paths that cover every topic entity and trace multi-hop relations across heterogeneous sources. These paths form interpretable chains of thought, revealing both answers and their cross-source support.

\myparagraphunderlinenew{To address multi-source verification}  
\srag computes a tri-factor score, combining source trustworthiness, cross-source corroboration, and entity-to-evidence alignment. Low-scoring branches are discarded before LLM calls, reducing token usage and preventing source-specific noise.


\myparagraphunderlinenew{To address multi-hop reasoning}
\srag generates an indicator in the question analysis stage that predicts the relationship depth between each topic entity and the answer. Guided by it, the system retrieves multi-hop paths from a predicted depth in the KG, enabling dynamic structured search. The same path requirement guides unstructured retrieval to connect related text chains across documents. Unlike approaches that restart retrieval at every step, \srag enhances LLMs to follow coherent reasoning paths that lead to the answer.

\myparagraphunderlinenew{To address multi-entity questions} 
\srag us-es a three-phase exploration process over the question subgraph, documents, and web results. All paths must include every topic entity in the order given by the skyline indicator. In structured retrieval, the paths are logical and faithful; in unstructured retrieval, keywords and their connections are searched across text. Each path yields one answer candidate and serves as an interpretable reasoning chain, leveraging both LLM and KG knowledge.

\myparagraphunderlinenew{To address graph-structure under-utilization}  
\srag 
forms a question subgraph by expanding topic entities to their maximal-depth neighbors and merging subgraphs from multiple KGs. We apply node clustering and graph reduction to cut the search costs and inject high-confidence text edges to dynamically fill KG gaps. During evidence exploration, a semantics-gated, multi-source-verified, bidirectional BFS prunes low-confidence branches early. 
Inspired by GoT~\cite{besta2024graphGoT}, \srag prompts the LLM to summarize the top-$W_{\max}$ paths before answer evaluation to further reduce hallucinations.
In summary, the advantages of \srag can be abbreviated as:


\noindent\textbf{Structured source-aware retrieval}: \srag integrates heterogeneous evidence from diverse sources into a unified structured representation, enabling seamless reasoning.



\noindent\textbf{Multi-source verification}: 
\srag prunes candidate paths based on both question relevance and cross-source corroboration before any LLM call, generating a compact, high-confidence context that reduces hallucinations and lowers LLM costs.

\noindent\textbf{Interpretable cross-source reasoning}: The extracted reasoning paths trace how facts from different modalities converge on the answer, providing transparent, step-by-step justification and enhancing the faithfulness of LLM outputs.

\noindent\textbf{Efficiency and adaptability}:  
a) \srag is a plug-and-play framework that can be seamlessly applied to various LLMs, KGs, and texts.
b) \srag is auto-refresh. New information is incorporated instantly via web retrieval instead of costly LLM fine-tuning.
c) \srag achieves state-of-the-art results on all the tested datasets, surpasses the strong hybrid baseline ToG-2 by an average of 20.3\% and up to 30.1\%, and enables smaller models to achieve reasoning performance comparable to GPT-4-Turbo.
\section{Related Work}
\label{sec:related_work}
\vspace{-2mm}
\myparagraph{Text-based RAG}  
Early text-based RAG systems embed queries and texts in a shared vector space and retrieve the closest chunks \cite{gao2023retrieval,wang2025llm,ding2025illusioncaptcha}. Iterative methods such as ITERRETGEN alternate between retrieval and generation to add context \cite{shao2023enhancing}, but coarse passages often mix relevant facts with noise, weakening the signal for reasoning. CoT prompts can guide retrieval toward deeper clues \cite{wei2022cot}, but they still rely on semantic similarity and ignore the structure of relations, so long-range connections may be missed or require many iterations to uncover.

\myparagraph{KG-based RAG}  
Graphs are widely used to model complex relationships among different entities \cite{sima2024deep, wang2024efficient, DBLP:journals/pvldb/WangWWZQZL24,wang2025effective,tan2023maximum,tan2023higher}.
KGs store triples, making entity links explicit \cite{hu2025mmapgtrainingfreeframeworkmultimodal,lifan2024adarisk, lifan2024hypergraph}. Agent-based methods let an LLM walk the graph hop by hop. ToG asks the LLM to choose the next neighbour at each step \cite{tog1.0sun2023think}, and StructGPT reformulates a structured query into repeated read-reason cycles \cite{jiang2023structgpt}. Plan-on-Graph and DoG run several LLM calls to rank candidate neighbours \cite{plan-on-graph,debated-on-graph}. But a walk starts from a single entity can miss answers that involve several topic entities and becomes fragile on long chains. Paths-over-Graph \cite{pogtan2025paths} focuses on multi-hop reasoning but relies solely on the KG, so it inherits KG gaps and rising update costs.



\myparagraph{Hybrid RAG}  
Recent work combines structured and unstructured sources. GraphRAG builds a document-level KG to guide passage retrieval \cite{graphragmicrosoft}, CoK mixes multiple sources to ground outputs \cite{li2023chaincok}, and HybridRAG unifies vector and KG retrieval in a single pipeline \cite{sarmah2024hybridrag}. Although these methods improve coverage, they retrieve each source separately and simply concatenate results, which can introduce redundant or low-quality evidence.
Agentic approaches like ReAct interleave reasoning with retrieval actions to reduce errors \cite{yao2022react}, but their modules still face the same coverage and granularity limitations. ToG-2 \cite{tog2.0ma2024think} queries all sources simultaneously, but it only retrieves one-hop neighbours and does not assess source reliability or cross-source consistency, making it unsuitable for multi-hop complex questions.


\section{Preliminary}\label{sec:prelim}

\begin{figure*}
    \centering
    \includegraphics[width=0.99 \linewidth]
    {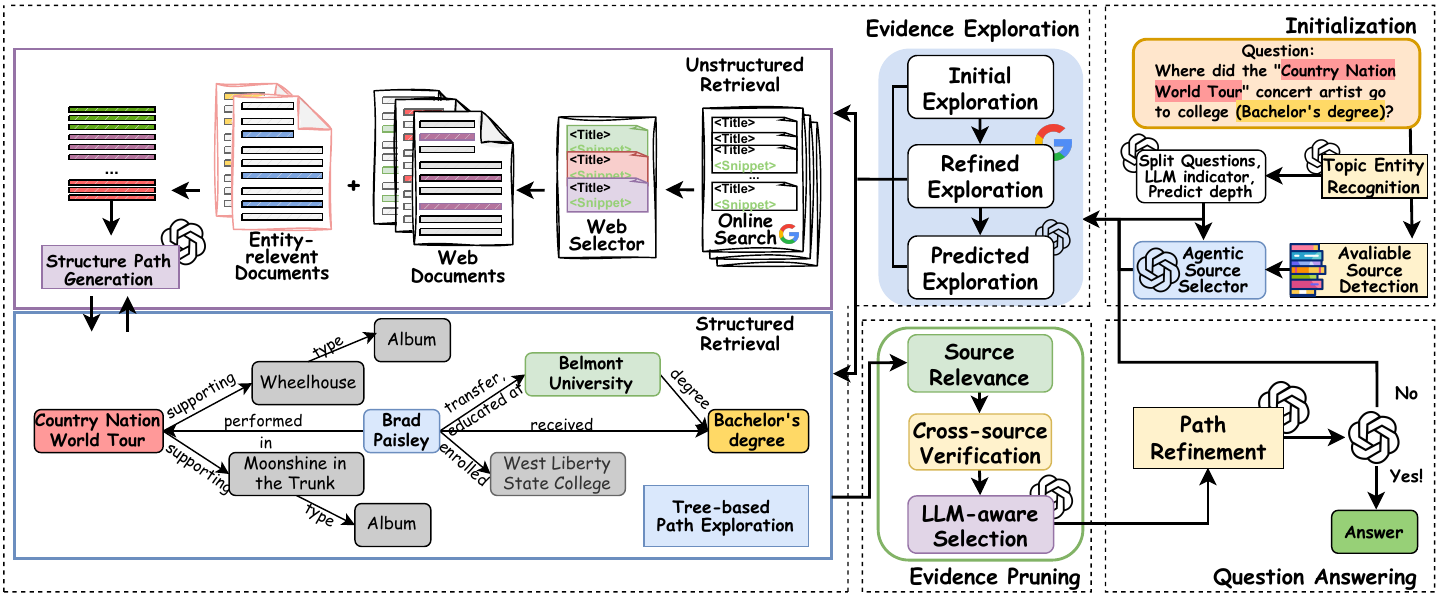}
    \caption{Overview of the \srag architecture. Evidence exploration: after initialization (detailed in Figure~\ref{fig:initial}), the model retrieves entity paths from diverse sources through three exploration phases. Evidence Pruning: \srag applies a three-step evidence pruning procedure after each exploration phase. Question Answering: the pruned paths are then evaluated for question answering. 
    }
    \vspace{-2mm}
    \label{fig:method}
\end{figure*}


Consider a Knowledge Graph (KG) $\mathcal{G(E,R)}$, where $\mathcal{E}$ and $\mathcal{R}$ represent the set of entities and relations, respectively. $\mathcal{G(E,R)}$ contains abundant factual knowledge in the form of triples, i.e., $\mathcal{G(E,R)}=\{(e_h, r, e_t)\mid e_h,e_t\in\mathcal{E}, r\in\mathcal{R}\}$.

\begin{definition}[Reasoning Path] 
Given a KG $\mathcal{G}$,
    a reasoning path within $\mathcal{G}$ is defined as a connected sequence of knowledge triples, represented as: ${\rm{path}}_{\mathcal{G}}(e_1, e_{l+1}) = \{(e_1,r_1,e_2),(e_2,r_2,e_3)$ $,...,(e_{l},r_l,e_{l+1})\}$, where $l$ denotes the length of the path, i.e., ${\rm{length}}({\rm{path}}_{\mathcal{G}}(e_1, e_{l+1})) = l$.

\end{definition}

\begin{definition}[Entity Path]
Given a KG $\mathcal{G}$ and an entity list $\text{list}_e$ = [$e_1, e_2, e_3, \ldots, e_l$], the entity path of $list_e$ is defined as a connected sequence of reasoning paths, which is denoted as ${\rm{path}}_{\mathcal{G}}(list_e)$
$= \{{\rm{path}}_{\mathcal{G}}(e_1, e_2), $ $
{\rm{path}}_{\mathcal{G}}(e_2$, $e_3), \ldots, {\rm{path}}_{\mathcal{G}}(e_{l-1}, e_l) \}=\{(e_s, r, e_t)$$| (e_s, r, e_t)$ $ \in {\rm{path}}_{\mathcal{G}}(e_{i}, e_{i+1}) $$\land 1 \leq i < l\}$.
\end{definition}

Knowledge Base Question Answering (KBQA)
is a fundamental reasoning task based on KBs. Given a natural language question $q$ and a KB $\mathcal{B}$, the objective is to devise a function $f$ that predicts answers $a \in \text{Answer}(q)$ utilizing knowledge encapsulated in $\mathcal{B}$, \ie $a = f(q, \mathcal{B})$.


\section{Method}\label{sec:method}
The \srag framework integrates multiple knowledge sources to ensure comprehensive and reliable retrieval. 
The overview of \srag is presented in Figure~\ref{fig:method}.
All sources are first detected and agentically selected in  Section~\ref{sec:med:initial}, and then fully retrieved and augmented in Section~\ref{sec:medthod:exploration}. These sources include three categories.
First, the \textbf{knowledge graph} provides the most accurate and structured evidence. 
For each question, we first extract an evidence subgraph $\mathcal{G}_q^{s}$ from every KG source (i.e., Freebase and WikiKG) and then merge these subgraphs into a single global evidence subgraph $\mathcal{G}_q$.
Second, \textbf{wikipedia documents} supply semi-structured information\footnote{\srag uses the Wikipedia page of each topic entity $e\in \mathcal{G}_{\text{WikiKG}}$ as the initial document.}. 
We retrieve question-relevant Wiki document set using the topic entity set $\text{Topic}(q)$, forming 
$
\text{Wiki} = \bigl\{\text{Doc}(e)\mid e\in\text{Topic}(q)\bigr\}.
$
Third, \textbf{web documents} capture real-time online results\footnote{\srag uses Google Search by SeripAPI for online retrieval.}. We issue an online search result set for $q$, yielding 
$
\mathrm{Web} = \text{OnlineSearch}(q),
$
where each search result includes a web page title, description snippet, and URL. The faithfulness of web evidence is later assessed in Section~\ref{sec:method:prun}.

\subsection{Step \RomanNumeralCaps{1}: Initialization}\label{sec:med:initial}
\begin{figure*}
    \centering
    \includegraphics[width=0.99 \linewidth]{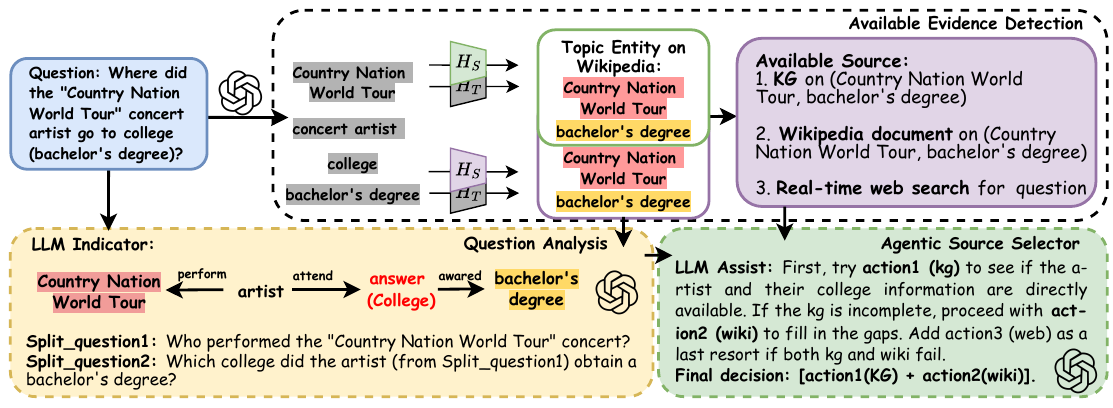}
    \caption{Overview of the initialization section. The initialization workflow diagram of three stages, i.e., available evidence detection, question analysis and agentic source selector.}
    \label{fig:initial}
\end{figure*}

The initialization has three main stages, i.e., available evidence detection, question analysis, and agentic source selector. The framework is shown in Figure~\ref{fig:initial}.


\myparagraph{Available evidence detection} Given a question $q$, \srag first identifies candidate KBs, including knowledge graphs, web pages, and documents. To determine which sources are relevant to $q$, \srag uses an LLM to extract potential topic entities. It then applies BERT-based similarity matching to align these entities with those in each source (e.g., $\mathcal{E}\in \{\mathcal{G}_{\text{freebase}}, \mathcal{G}_{\text{WikiKG}}\}$). As shown in Figure~\ref{fig:initial}, we encode the extracted entities and all entities from a source into dense embeddings $H_{T}$ and $H_{\mathcal{S}}$, and compute a cosine similarity matrix to identify matches. For each extracted entity and each knowledge source, entities whose similarity exceeds a threshold form the set $\mathrm{Topic}(q)$. Each source maintains its own $\mathrm{Topic}(q)$; if $\lvert \mathrm{Topic}(q)\rvert>0$, the source is marked relevant and added to the total sources list $S_t \subseteq \{\text{KG, Wiki, Web}\}$ for further agentic selection. The $S_t=\{\text{Web}\}$ is considered as the initial setting. This set underlies the construction of the question-related subgraph and the preparation of documents in later steps.

\myparagraph{Question analysis} To reduce hallucinations, the question analysis phase is divided into two parts and executed within a single LLM call using an example-based prompt (detailed in Appendix~\ref{appendix:prompt}). 
First, it breaks the complex question $q$ into sub-questions, each linking one topic entity to the potential answer; solving these sub-questions together grounds the original query.  
Second, a solving skyline is generated, which lists all topic entities and predicts the answer’s position in a single chain of thought derived from $q$. This skyline captures the relationships and order among the entities and the answer, transforming the complex question into a concise, simplified reasoning path. From this, we compute a predicted depth $D_{\text{predict}}$, defined as the maximum distance between the predicted answer and any topic entity. An example of question analysis, with $D_{\text{predict}}=2$, is shown in Figure~\ref{fig:initial}.

\myparagraph{Agentic source selector}
Most existing systems operate on a single KG or KB.  
Hybrid RAG methods~\cite{tog2.0ma2024think, li2023chaincok} can combine multiple information sources, but they typically query a fixed set (usually one or two) and ignore the question‑specific trade‑off between coverage and cost.  
Blindly querying every possible source greatly increases latency and computation.

To address this limitation, we introduce an agentic source selector.  
Given the total evidence source list $S_t$ and question analysis result, an LLM-selected agent analyses the incoming question and chooses an initial source combination $S_a$ that best balances three factors:  
(i) time sensitivity,  
(ii) reasoning complexity, and  
(iii) domain relevance.  
Only the selected sources $S_a \subseteq S_t$ are used in the initial exploration stage in Section~\ref{method:Initial_Exploration}, reducing cost while preserving answer quality.

\subsection{Step \RomanNumeralCaps{2}:  Evidence Exploration}
\label{sec:medthod:exploration}

As discussed in Section~\ref{sec:intro}, finding reasoning paths that include all topic entities is essential for deriving accurate answers. These paths act as \textbf{interpretable chains of thoughts}, showing both the answer and the inference steps leading to it.

However, the evidence needed to complete such paths is often distributed across sources. 
Combining these heterogeneous sources is therefore as important as path-finding itself.
To discover high-quality paths while unifying evidence in a common format, the exploration is divided into three phases: initial exploration, refined exploration, and predicted exploration. 
In each exploration, retrievals from different sources are processed in parallel; After each phase, we apply path pruning and attempt to answer the question.
If a valid path is found, the search terminates; otherwise, it proceeds to the next phase. Due to space constraints, the pseudo-code for exploration is provided in Appendix~\ref{appendix:sec:explora}.

\subsubsection{Initial Exploration}
\label{method:Initial_Exploration}

To reduce LLM usage and narrow the search space, \srag first explores agent-selected knowledge sources in parallel. Structured and unstructured inputs are processed independently: structured retrieval captures explicit relational facts, whereas unstructured retrieval supports more complex or implicit reasoning.


\myparagraph{Structured retrieval}
For structured retrieval, we first detect an evidence subgraph from KGs, then explore topic-entity paths.

\myparagraphunderline{Subgraph detection}  
Inspired by \cite{pogtan2025paths}, we construct a $D_{\max}$-hop global evidence subgraph $\mathcal{G}_q$. For each topic entity, we retrieve all triples involving its $D_{\max}$-hop neighbors to incorporate relevant and faithful KG information into $\mathcal{G}^s_q$ from each knowledge source, i.e., $s\in\{$Freebase, WikiKG$\}$. 
To enhance knowledge coverage, we also merge multiple $\mathcal{G}^s_q$ into a global graph $\mathcal{G}_q$. To control information overload and reduce computation, we apply node and relation clustering, along with graph reduction techniques, to prune $\mathcal{G}_q$ effectively.

\myparagraphunderline{Tree‑based path retrieval}  
Instead of the maximum depth $D_{\max}$, \srag performs initial exploration at the predicted depth $D_{\text{predict}}$.
Given the subgraph $\mathcal{G}_q$, the ordered topic entity set $\mathrm{Topic}(q)$, the skyline indicator $I_{\text{sky}}$, and the depth $D = \min(D_{\text{predict}}, D_{\max})$, we identify candidate reasoning paths that include all topic entities in order. To avoid exhaustive search, we apply a tree-structured  bidirectional breadth-first search (BiBFS) from each topic entity to extract a set of all potential entity paths, defined as:
Paths$_I = \{p \mid |\text{Topic}(q)| \cdot (D{-}1) < \operatorname{length}(p) \leq |\text{Topic}(q)| \cdot D\}$.

At each step, a \textbf{cross-score} (introduced in Section~\ref{sec:method:prun}) is computed between the path, the skyline indicator, and retrieved documents to prune unpromising branches. Only the top-$W_1$ paths are retained as seeds for further expansion. This method enables efficient construction of high-quality candidate paths while maintaining interpretability. The pseudo-code for structured retrieval is detailed in Algorithm~\ref{algorithm:StructuredRetrieval} of Appendix~\ref{appendix:sec:explora}.


\myparagraph{Unstructured retrieval}  
For each document $\text{Doc}(e)$ associated with $e \in \mathrm{Topic}(q)$, we retrieve text blocks, split them into smaller passages, and select the top-$W_{\max}$ sentences using a dense retrieval model (DRM). Instead of embedding the full query, \srag   uses the skyline indicator to emphasize structural relevance. Unlike ToG-2.0, which targets only one-hop relations, ours captures more complex reasoning, i.e., transitive multi-hop relations.
The resulting sentences are used to prompt the LLM to construct new knowledge paths, which are summarized and added to $\text{Paths}_I$. 


\myparagraphunderline{Web document retrieval}  
When offline documents and KGs are insufficient, \srag performs online retrieval by issuing the question $q$ to a search engine and prompting the LLM to select the top-$W_{\max}$ web results. These documents are then processed using the same DRM-based screening and path construction as in the offline setting.
The pseudo-code and prompting for unstructured retrieval are detailed in Algorithm~\ref{algorithm:UnstructuredRetrieval} of  Appendix~\ref{appendix:sec:explora} and Appendix~\ref{appendix:prompt}.

By combining KG-based, document-based, and web-based retrieval, \srag generates a rich and interpretable path set as evidence,  which is passed to the subsequent pruning stages.


\subsubsection{Refined Exploration}
Traditional KG-based reasoning typically reuses stored facts through a complex retrieval process. However, this approach often falls into fast-evolving or emerging information, which may not be adequately represented in the KG. To overcome this limitation, \srag introduces a novel mechanism that leverages the LLM’s ability to generate follow-up questions and refine the knowledge search. 
Specifically, \srag prompts the LLM to generate a follow-up question, $q_{\text{new}}$, along with a new skyline indicator, $I_{\text{new}}$, which signals the additional information required beyond what is currently represented in the knowledge graph.
The follow-up question $q_{\text{new}}$ is designed to explicitly target the new information or emerging concepts, ensuring that the retrieval process captures relevant, up-to-date data. 
From this exploration, all the available knowledge sources $S_t$ will be utilized for retrieval.
Using $q_{\text{new}}$, we extract $\mathrm{Topic}(q_{\text{new}})$ and perform unstructured retrieval: both new and historical documents are ranked according to $I_{\text{new}}$. For structured retrieval, we set the search depth $D = D_{\max}$ and use $I_{\text{new}}$ to guide exploration within the KG. All paths retrieved in this phase are added to the refined entity path set $\mathrm{Paths}_{R}$ for further pruning.

\subsubsection{Predict Exploration}

In many RAG systems, LLMs merely rephrase facts rather than leveraging their own implicit knowledge. To address this, \srag   encourages LLMs to generate predictions using their path understanding and implicit knowledge, offering additional valuable insights. This involves creating new skyline indicators, $I_{\text{Pred}}$, for the predicted entities, $e \in \text{Predict}(q)$, and using text similarity to confirm and align them with $\mathcal{E}q \in \mathcal{G}q$. An entity list, $\text{List}_{P}(e) = \text{Topic}(q) + e$, is formed and ranked based on $I_{\text{pred}}$ to enhance reasoning effectiveness. 

For structured retrieval, predicted entity paths $\text{Paths}_{P}$ are extracted from $\mathcal{G}_q$ at a fixed depth $D_{\max}$:
$
\text{Paths}_{P} = \{ p \mid \text{length}(p) \leq |\text{Topic}(q)| \cdot D_{\max} \},
$
where $p = \text{Path}_{\mathcal{G}_q}(\text{List}_{P}(e))$. 
For unstructured retrieval, the pair $(q, I_{\text{pred}(q)})$ is used to retrieve and score relevant sentences. The resulting paths are added to $\text{Paths}_{P}$.
These paths with new indicators
are evaluated similarly to the initial exploration and refined exploration phases. The prompting template is shown in Appendix~\ref{appendix:prompt}.

\subsection{Step \RomanNumeralCaps{3}: Evidence Pruning}
\label{sec:method:prun}

\myparagraph{Multi-source verification in pruning}  
Traditional LLM–QA pipelines typically perform two-step pruning: an embedding filter narrows down the candidate set, followed by an LLM agent that selects the most relevant evidence. However, this method assumes uniform evidence sources. When the corpus includes diverse modalities, such as structured knowledge graphs, semi-structured Wiki pages, and unstructured web content, pruning solely by relevance can either discard crucial facts or retain redundant information (over- or under-pruning).

The \srag addresses this by adding a multi-source verification term to the relevance score. This term up-weights paths that are corroborated across heterogeneous sources and down-weights isolated claims from less reliable modalities. As a result, pruning balances topic relevance with cross-modal agreement, producing a compact yet reliable evidence set for downstream reasoning\footnote {This module is model‑agnostic; we demonstrate it with \srag , but it can be inserted into any KG+RAG pipeline.}.
Due to space constraints, the pseudo-code for evidence pruning is summarized in Algorithm~\ref{algorithm:EvidencePruning} of Appendix~\ref{appendix:sec:pruning}.


Formally, let $\mathcal{C}=\{p_i\}_{i=1}^{N}$ as the candidate evidence paths, each associated with three scores $(s^{\text{rel}}_i,\; s^{\text{ver}}_i,\; s^{\text{llm}}_i) \in [0,1]^3$ denoting relevance, verification, and LLM compatibility, respectively. The derivation of each score is described below.

\myparagraph{Source relevance}
\label{sec:prune_relevance}
Given a query skyline indicator $I$ and its topic‑entity set $\mathrm{Topic}(q)$, we compute a hybrid relevance score:
\vspace{-1mm}
\[
\begin{aligned}
s^{\text{rel}}_i
  &=
  \lambda_\text{sem}\!\cdot\!
  \underbrace{\cos\bigl(\mathbf{h}(I),\mathbf{h}({p_i})\bigr)}_{\text{semantic}}
  \\[2pt]
  &\quad+\;
  \lambda_\text{ent}\!\cdot\!
  \underbrace{\mathrm{Jaccard}\bigl(\mathrm{Topic}(q),\mathrm{Ent}(p_i)\bigr)}_{\text{entity\,overlap}},
\end{aligned}
\]
\vspace{-1mm}
\noindent where $\mathbf{h}{(\cdot)}$ denotes sentence‑level embeddings by DRM,  
$\mathrm{Ent}(p_i)$ extracts linked entities in $p_i$, and $\lambda_\text{sem} + \lambda_\text{ent}=1$.  
The top-$W_1$ paths form a candidate pool $\widetilde{\mathcal{C}} $ for cross-source evaluation.

\myparagraph{Cross‑source verification}
\label{sec:prune_verification}
We estimate the reliability of each candidate path using three reliability features: (i) source reliability, (ii) corroboration from independent sources, and (iii) consistency with existing KG facts. Candidates in $\widetilde{\mathcal{C}}$ are grouped by provenance into $\mathcal{C}_{\text{KG}}$, $\mathcal{C}_{\text{Wiki}}$, and $\mathcal{C}_{\text{Web}}$. For each path $p_i$, the supporting external sources are:
$
\operatorname{Supp}(p_i)=\{\operatorname{src}(p_j)\mid \mathrm{Sim}(p_i,p_j)\ge\gamma\},
$
where $\operatorname{src}(\cdot)$ returns the source type, and $\gamma$ is a cosine similarity threshold. 
The reliability features inside are defined as:
\vspace{-1mm}
\[
\begin{aligned}
f_1(p_i) &= \rho_{\mathrm{src}(p_i)}, \quad \rho_{\mathrm{KG}} > \rho_{\mathrm{Wiki}} > \rho_{\mathrm{Web}}, \\[4pt]
f_2(p_i) &= \frac{\min(|\operatorname{Supp}(p_i)|, W_{\text{max}})}{W_{\text{max}}}, \\[4pt]
f_3(p_i) &= \frac{|\mathrm{Ent}(p_i) \cap \mathcal{E}_q|}{|\mathrm{Ent}(p_i)|}, \quad \mathcal{E}_q \in \mathcal{G}_q. 
\end{aligned}
\vspace{-1mm}
\]


\noindent 
The verification score is computed as $s^{\text{ver}}_{i} = \sum_{k=1}^{3}\alpha_k\,$ $f_k(p_i)$,
where coefficients $\alpha_k$ are non-negative and $\sum_k \alpha_k = 1$. Each candidate path in $\widetilde{\mathcal{C}}$ is then ranked by a \textbf{cross-score} that combines relevance and verification:
$
\text{cross-score}(p_i) = \alpha_{\text{cross}} \cdot s_i^{\text{rel}} + (1-\alpha_{\text{cross}}) \cdot s_i^{\text{ver}}.
$
\noindent The top-$W_2$ paths are selected for the final LLM-driven pruning.

\myparagraph{LLM‑aware selection}
At this stage, we prompt the LLM to score and select the top-$W_{\max}$ reasoning paths most likely to contain the correct answer. The specific prompt used to guide LLM in the selection phase can be found in Appendix~\ref{appendix:prompt}.






\subsection{Step \RomanNumeralCaps{4}: Question Answering}
\vspace{-1mm}

Utilizing the pruned paths obtained in Section~\ref{sec:method:prun}, we propose a two-step question-answering strategy, emphasizing deep thinking and slow reasoning.

\myparagraph{Path Refinement}
To ensure accurate reasoning and mitigate hallucinations, we prompt LLMs to refine the provided paths. 
By evaluating and selecting only relevant facts, the paths are summarized into concise, focused evidence, suitable for subsequent reasoning. Prompt details are in Appendix~\ref{appendix:prompt}.


\myparagraph{CoT Answering}
Following path refinement, the LLM employs a CoT prompting method to reason  systematically through the refined evidence paths.  It first checks whether they answer each subquestion and the full question. If the evaluation is positive, LLM generates the answer using the paths, along with the question and question analysis results as inputs, as shown in Figures~\ref{fig:method}. The prompts for evaluation and generation are in Appendix~\ref{appendix:prompt}. 
If negative, another exploration round begins. When all rounds end without a valid answer, the LLM replies using the given paths and its inherent knowledge. Additional details on the prompts can be found in Appendix~\ref{appendix:prompt}.
\section{Experiment}
\label{sec:experiment}
\vspace{-2mm}




In this section, we evaluate \srag on seven benchmark KBQA datasets. Besides the \srag proposed in this paper, we introduce \textbf{\srag-E}, which randomly selects one relation from each edge in the clustered question subgraph to evaluate the impact of graph structure on KG involved LLM reasoning.  The detailed experimental settings, including datasets, baselines, and implementations, can be found in Appendix~\ref{appen:dataset_details}.

\begin{table*}
    \centering

\caption{Results of \srag across all datasets, compared with the state-of-the-art (SOTA) with GPT-3.5-Turbo. The highest scores are highlighted in bold, while the second-best results are underlined for each dataset.}
\label{figure:mainresult}
\setlength\tabcolsep{1.0 pt}

\resizebox{1\linewidth}{!}{
\begin{tabular}{lccccccccc}
\toprule
\multirow{2}{*}{\textbf{Type}} & \multirow{2}{*}{\textbf{Method}} & \multirow{2}{*}{\textbf{LLM}} & \multicolumn{4}{c}{\textbf{Multi-Hop KBQA}} & \textbf{Single-Hop KBQA} & \textbf{Slot Filling} & \textbf{Open-Domain QA} \\ 
\cmidrule(r){4-7}\cmidrule(r){8-8}\cmidrule(r){9-9}\cmidrule(r){10-10}
 &  &  & CWQ & WebQSP & AdvHotpotQA & QALD10-en & SimpleQA & ZeroShot RE & WebQuestions \\ 
\toprule
\multirow{3}{*}{LLM-only} 
    & IO prompt & \multirow{3}{*}{GPT-3.5-Turbo} & 37.6 & 63.3 & 23.1 & 42.0 & 20.0 & 27.7 & 48.7 \\
    & CoT \cite{wei2022cot}       &                         & 38.8 & 62.2 & 30.8 & 42.9 & 20.3 & 28.8 & 48.5 \\
    & SC  \cite{wang2022self}      &                         & 45.4 & 61.1 & 34.4 & 45.3 & 18.9 & 45.4 & 50.3 \\ 
\toprule
\multirow{2}{*}{Vanilla RAG} 
    & Web-based  & \multirow{2}{*}{GPT-3.5-Turbo} & 41.2 & 56.8 & 28.9 & 36.0   & 26.9 & 62.2 & 46.8 \\
    & Text-based &                         & 33.8 & 67.9 & 23.7 & 42.4 & 21.4 & 29.5 & 35.8 \\ 
\toprule
\multirow{3}{*}{KG-based RAG} 
    & ToG \cite{tog1.0sun2023think} & GPT-3.5-Turbo & 58.9 & 76.2 & 26.3 & 50.2 & 53.6 & 88.0 & 54.5 \\
    & ToG \cite{tog1.0sun2023think} & GPT-4   & 69.5 & 82.6 & -    & 54.7 & 66.7 & 88.3 & 57.9 \\
    & PoG \cite{pogtan2025paths} & GPT-3.5-Turbo & 74.7 & 93.9 & -    & -    & 80.8 & -    & 81.8 \\ 
\toprule
\multirow{2}{*}{Hybrid RAG} 
    & CoK \cite{li2023chaincok}   & \multirow{2}{*}{GPT-3.5-Turbo} & -   & 77.6 & 35.4 & 47.1 & -   & 75.5 & -   \\
    & ToG-2 \cite{tog2.0ma2024think} &                         & -   & 81.1 & 42.9 & 54.1 & -   & 91.0 & -   \\ 
\toprule
\multirow{4}{*}{Proposed} 
    & \srag-E  & \multirow{2}{*}{Llama-3.1-70B} & 71.3 & 89.7 & 48.4 & 70.9 & 80.4 & 95.6 & 76.8 \\
    & \srag     &                               & 75.6 & 93.0 & {\ul 55.2} & 76.0 & {\ul 85.9} & 94.2 & 81.4 \\ 
\cline{3-10}
    & \srag-E  & \multirow{2}{*}{GPT-3.5-Turbo} & {\ul 76.8} & {\ul 94.0} & 51.3 & {\ul 81.1} & 81.7 & {\ul 96.9} & {\ul 85.2} \\
    & \srag     &                         & \textbf{81.2} & \textbf{96.1} & \textbf{58.9} & \textbf{84.2} & \textbf{88.8} & \textbf{97.7} & \textbf{88.3} \\ 
\toprule

\end{tabular}
\vspace{-2mm}

}
\vspace{-2mm}

\end{table*}

\subsection{Main Results}
Since \srag leverages external knowledge, we first compare it against other RAG-based methods. As shown in Table~\ref{figure:mainresult}, \srag achieves SOTA results across all datasets, outperforming prior SOTA by an average of 10.8\% and up to 30.1\% on QALD10-en. Compared with ToG-2, a strong hybrid RAG baseline, \srag achieves average improvements of 20.3\%, up to 30.1\% on QALD10-en. 
Against Llama3.1-70B, a weaker reasoning model, \srag shows an average gain of 9.2\% on 5 datasets, up to 21.9\% on QALD10-en  compared to previous GPT-3.5-based methods, and even surpasses the powerful GPT-4-based ToG baseline by 14. 4\% on average, up to 23.5\% on WebQuestions. This indicates \srag   significantly enhances the reasoning abilities of less powerful LLMs by providing faithful and interpretable cross-source knowledge paths. 
Additionally, compared to vanilla text/web-based RAG methods, \srag   shows average gains of 45.5\%, up to 68.2\% on ZeroShot RE. 

When compared to methods without external knowledge (IO, CoT, SC), \srag   improves accuracy by 41.5\%  on average, up to 68.5\% on SimpleQA. 
 Notably, while vanilla RAG methods and LLM-only approaches show similar performance due to overlapping training corpora, \srag achieves superior results using the almost same corpus, highlighting its advanced unstructured retrieval capability.
The variant \srag-E also surpasses existing SOTA methods by 6.8\% on average, up to 24.0\% on QALD10-en. 
These findings demonstrate \srag is excellent for reasoning tasks, particularly for complex logical reasoning. By retrieving deeply and integrating the structural information of the question from diverse knowledge sources, it enhances the deep reasoning capabilities of LLMs, leading to superior performance.

\subsection{Ablation Study}\label{exp:ablation}


\begin{table*}
    \centering
    \caption{Performance of the IO baseline and \srag across four datasets on different backbone models. The highest improvement is highlighted in bold, while the second-best results are underlined for each model.}
    \label{fig:different_backbone}
    \resizebox{1\textwidth}{!}{%
    \begin{tabular}{l
                    ccc ccc ccc ccc ccc}
        \toprule
        \multirow{2}{*}{\textbf{Dataset}} &
        \multicolumn{3}{c}{\textbf{Llama-3.1-8B}} &
        \multicolumn{3}{c}{\textbf{Llama-3.1-70B}} &
        \multicolumn{3}{c}{\textbf{DeepSeek-v3}} &
        \multicolumn{3}{c}{\textbf{GPT-3.5-Turbo}} &
        \multicolumn{3}{c}{\textbf{GPT-4-Turbo}} \\[2pt]
        \cmidrule(lr){2-4}\cmidrule(lr){5-7}\cmidrule(lr){8-10}\cmidrule(lr){11-13}\cmidrule(lr){14-16}
        & IO & \srag   & \%$\uparrow$ 
        & IO & \srag   & \%$\uparrow$
        & IO & \srag   & \%$\uparrow$
        & IO & \srag   & \%$\uparrow$
        & IO & \srag   & \%$\uparrow$ \\
        \midrule
        AdvHotpotQA  & 16.9 & 35.6 &    111   & 21.7 & 48.4 & 123       & 27.8 & 55.4 &  99.0         & 23.1 & 56.2 & {\ul143} & 46.4 & 67.9 &  {\ul46.0} \\
        WebQSP       & 38.5 & 86.0 &    {\ul 123}         & 56.2 & 95.2 &  69.0       & 68.0 & 97.7 &  44.0         & 66.3 & 96.9 &  46.0 & 75.4  & 98.2  &  30.0 \\
        CWQ          & 29.8 & 62.4 &    109         & 35.4 & 83.2 & {\ul135}    & 38.7 & 84.5 & {\ul118}    & 39.2 & 84.0 & 114 & 45.3  & 89.7  &  \textbf{98.0} \\
        ZeroShot RE & 27.2 & 77.5 & \textbf{185}   & 34.6 & 97.5 & \textbf{182} & 38.6 & 97.0 & \textbf{151} & 37.2 & 97.7 & \textbf{163} & 49.8  & 98.5  &  \textbf{98.0} \\
        \bottomrule
    \end{tabular}}
\end{table*}

\myparagraphquestion{How does the effectiveness of \srag   vary with different LLM capabilities}
We evaluated \srag with five LLM backbones (LLama-3.1-8B, Llama-3.1-70B, Deepseek-v3, GPT-3.5-Turbo, GPT-4-Turbo) on three multi-hop datasets (AdvHotpotQA, WebQSP, CWQ) and one slot-filling dataset (ZeroShot RE). As shown in Table~\ref{fig:different_backbone}, \srag improves performance across all models and datasets by an average of 109\%. Notably, it boosts Llama-3.1-8B by 132\% on average, up to 185\% on ZeroShot RE. This brings weaker models close to and even surpasses the direct reasoning accuracy of GPT-4-Turbo, confirming that \srag alleviates knowledge and comprehension bottlenecks. Stronger models also benefit from \srag. 
GPT-3.5-Turbo and GPT-4-Turbo are improved on complex reasoning tasks, although the improvement decreases slightly as their inherent reasoning is already strong. 
Even so, \srag yields a 98\% improvement on CWQ and ZeroShot RE with the most capable LLMs. Overall, \srag enables deeper knowledge retrieval and more reliable and interpretable reasoning across LLMs of varying strength, rather than relying solely on their inherent knowledge.

\textit{To further evaluate the performance of \srag,
we conduct additional ablation studies on search depth,  agentic source selector, prompt setting, and knowledge sources. The detailed results are shown in Appendix~\ref{appendix:Additional_Ablation}.}

\subsection{Effectiveness Evaluation}\label{exp:result_analysis}

\begin{figure}
    \centering
    \includegraphics[width=1\linewidth]{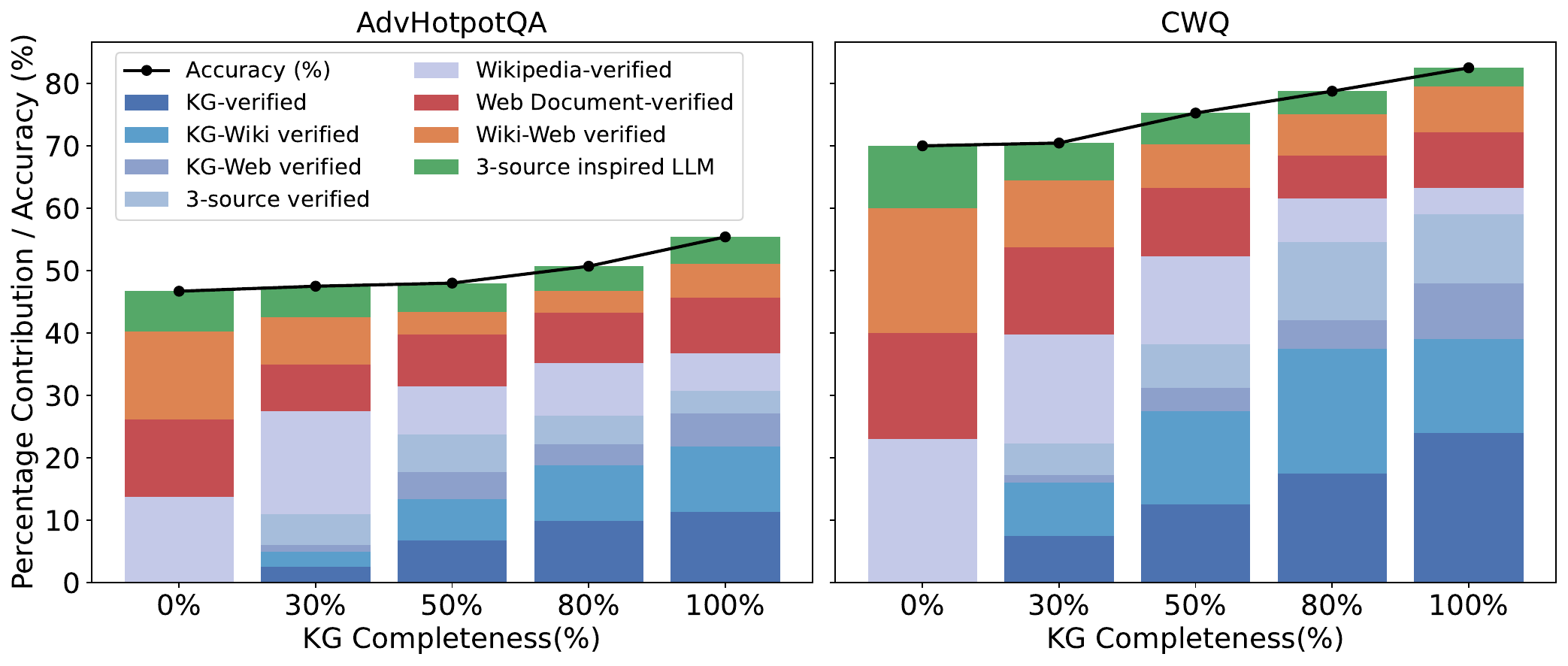}
    \caption{Accuracy and answer source composition by varying KG completeness on AdvHotpotQA and CWQ.}\label{fig:Source_composition_icmpleKG}
    \label{fig:enter-label}
\end{figure}

\myparagraph{Effectiveness on incomplete KG}
To evaluate how \srag addresses KG incompleteness and the impact of graph quality on reasoning performance, we constructed KGs with varying completeness levels (0\%, 30\%, 50\%, 80\%, and 100\%) on the AdvHotpotQA and CWQ. For each completeness level, we randomly selected a corresponding proportion of triples to build a new KG, with the remainder removed. Results in Figure~\ref{fig:Source_composition_icmpleKG} indicate that accuracy decreases slightly, rather than dramatically, as incompleteness increases. To investigate this trend, we analyze contributions from different KG completeness levels, with detailed analyses presented in Appendix~\ref{appendix:fathhful_analysis}. The analysis reveals that at lower KG completeness, answers predominantly rely on wiki and web documents; as completeness increases, KG-based answers become dominant. This demonstrates that \srag does not solely depend on KG data and effectively mitigates KG incompleteness issues, highlighting its adaptability.

\textit{To further evaluate the performance, we perform additional experiments, including additional effectiveness evaluation on cross-source verification, multi-hop reasoning, multi-entity questions, and graph structure pruning in Appendix~\ref{appendix:Additional_Effectiveness}; reasoning faithfulness analysis in Appendix~\ref{appendix:fathhful_analysis}; error analysis in Appendix~\ref{appendix:exp:error_analysis}; efficiency analysis in Appendix~\ref{appendix:effiency_analysis}; and case study on cross-verified interpretable reasoning in Appendix~\ref{castudy}. A detailed outline is shown in \hyperref[outline]{Appendix Outline}.}

\vspace{-1mm}
\section{Conclusion}
\label{sec:conclusion}
\vspace{-1mm}

In this work, 
we introduce \srag, a structured source-aware retrieval method 
for faithful and transparent LLM reasoning. 
\srag answers complex questions with agent-driven, structured and unstructured,  multi-hop evidence exploration, ensuring every topic entity is linked across all knowledge corpora.
Efficiency is enhanced by a tri-factor cross-source verification, scoring, and early pruning discards low-quality branches before any generation step. 
Extensive experiments on 7 datasets show that \srag outperforms existing baselines, showcasing its superior reasoning capabilities and interoperability. 

\section{Ethics Statement}
In this work, we employ LLMs as the final selector through LLM-aware selection, rather than for open-ended text generation. As a result, the ethical risks associated with our method are expected to be lower than using LLMs for text generation. However, recent studies indicate that CoT prompting may introduce ethical biases \cite{shaikh-etal-2023-second}. Additionally, integrating evidence from multiple retrieval sources may also introduce or amplify ethical biases. In future work, we plan to systematically investigate the manifestation and impact of these biases in our method.

\section{Limitation}
The primary limitation of our proposed \srag framework is its exclusive focus on character-based knowledge sources. \srag does not incorporate external modalities such as images or videos, which can also contain substantial factual information. Integrating visual sources alongside textual evidence remains an important direction for future work and could further enhance the reasoning capabilities of the framework.

\section{Acknowledgment}
Xiaoyang Wang is supported by the Australian Research Council
DP230101445 and DP240101322. Wenjie Zhang is supported by the
Australian Research Council DP230101445 and FT210100303.
\bibliography{custom}
\appendix
\newpage

\label{sec:appendix}


\newpage
\onecolumn

\section*{Appendix Outline}
\label{outline}

\hypersetup{hidelinks}
\noindent
A.\,\hyperref[appendix:algorithm]{Algorithm}\dotfill\pageref{appendix:algorithm}\\
\hspace*{1.5em}A.1\,\hyperref[appendix:sec:explora]{Exploration}\dotfill\pageref{appendix:sec:explora}\\
\hspace*{3em}Algorithm~\ref{algorithm:StructuredRetrieval}: Structured\_Retrieval\dotfill\pageref{algorithm:StructuredRetrieval}\\
\hspace*{3em}Algorithm~\ref{algorithm:UnstructuredRetrieval}: Unstructured\_Retrieval\dotfill\pageref{algorithm:UnstructuredRetrieval}\\
\hspace*{3em}Algorithm~\ref{algorithm:EvidenceExploration}: Evidence\_Exploration\dotfill\pageref{algorithm:EvidenceExploration}\\[3pt]
\hspace*{1.5em}A.2\,\hyperref[appendix:sec:pruning]{Evidence pruning}\dotfill\pageref{appendix:sec:pruning}\\
\hspace*{3em}Algorithm~\ref{algorithm:EvidencePruning}: Evidence\_Pruning\dotfill\pageref{algorithm:EvidencePruning}\\[5pt]
\noindent
B.\,\hyperref[appendix:Additional_expriment]{Experiment}\dotfill\pageref{appendix:Additional_expriment}\\
\hspace*{1.5em}B.1\,\hyperref[appendix:Additional_Ablation]{Additional Ablation Study}\dotfill\pageref{appendix:Additional_Ablation}\\
\hspace*{3em}Does search depth matter?\dotfill\pageref{exp:search_depth}\\
\hspace*{3em}How does the agentic source selector affect performance?\dotfill\pageref{exp:agentic_source}\\
\hspace*{3em}How do path refinement prompts affect performance? \dotfill\pageref{exp:path refinement}\\
\hspace*{3em}How do different knowledge sources affect the performance of \srag?\dotfill\pageref{exp:diifferent knowledge sources affect the perfor}\\[3pt]
\hspace*{1.5em}B.2\,\hyperref[appendix:Additional_Effectiveness]{Additional Effectiveness Evaluation}\dotfill\pageref{appendix:Additional_Effectiveness}\\
\hspace*{3em}Effectiveness on cross-source verification.\dotfill\pageref{exp:cross-source_verification}\\
\hspace*{3em}Effectiveness on multi-hop reasoning.\dotfill\pageref{exp:multi-hop}\\
\hspace*{3em}Effectiveness on multi-entity questions.\dotfill\pageref{exp:multi-entity}\\
\hspace*{3em}Effectiveness on graph structure pruning.\dotfill\pageref{exp:structure_pruning}\\[3pt]
\hspace*{1.5em}B.3\,\hyperref[appendix:fathhful_analysis]{Reasoning Faithfulness Analysis}\dotfill\pageref{appendix:fathhful_analysis}\\
\hspace*{3em}Evidence of answer exploration sources\dotfill\pageref{exp:Exploration Source}\\
\hspace*{3em}Overlap ratio between explored paths and ground-truth paths\dotfill\pageref{exp:Overlapground-truth paths}\\[3pt]
\hspace*{1.5em}B.4\,\hyperref[appendix:exp:error_analysis]{Error Analysis}\dotfill\pageref{appendix:exp:error_analysis}\\[3pt]
\hspace*{1.5em}B.5\,\hyperref[appendix:effiency_analysis]{Efficiency Analysis}\dotfill\pageref{appendix:effiency_analysis}\\
\hspace*{3em}LLM calls cost analysis\dotfill\pageref{exp:LLM calls cost}\\
\hspace*{3em}Efficiency analysis on AdvHotpotQA\dotfill\pageref{exp:Efficiency Analysis on AdvHotpotQA}\\
[3pt]
\noindent
C.\,\hyperref[appen:dataset_details]{Experiment Details}\dotfill\pageref{appen:dataset_details}\\
\hspace*{3em}Experiment datasets\dotfill\pageref{fig:appendix:dataset}\\
\hspace*{3em}Experiment baselines\dotfill\pageref{exp:Baselines}\\
\hspace*{3em}Experiment implementation\dotfill\pageref{exp:implementation}\\
[3pt]
\noindent
D.\,\hyperref[castudy]{Case Study: Multi-Source Cross-Verified Interpretable Reasoning.}\dotfill\pageref{castudy}\\
\hspace*{3em}Case study example on KG-Wikipedia cross-verified reasoning\dotfill\pageref{tab:casestudy-kg-wiki}\\
\hspace*{3em}Case study example on KG-Web cross-verified  reasoning\dotfill\pageref{tab:casestudy-KG-web}\\
\hspace*{3em}Case study example on all source cross-verified reasoning (\RomanNumeralCaps{1})\dotfill\pageref{tab:3-soucr}\\
\hspace*{3em}Case study example on all source cross-verified reasoning (\RomanNumeralCaps{2})\dotfill\pageref{tab:casestudy-3-souces2}\\[3pt]
\noindent
E.\,\hyperref[appendix:prompt]{Prompts}\dotfill\pageref{appendix:prompt}\\
\hspace*{3em}Question analysis prompt template\dotfill\pageref{prompt:Question Analysis}\\
\hspace*{3em}Agentic source selector prompt template\dotfill\pageref{prompt:AgenticSourceSelector}\\
\hspace*{3em}From paragraph to knowledge path prompt template\dotfill\pageref{prompt:ParagraphtoKnowledge}\\
\hspace*{3em}Refined exploration prompt template\dotfill\pageref{prmpt:prompts_Refined_Exploration}\\
\hspace*{3em}Predict exploration prompt template\dotfill\pageref{prompt:Predict_Exploration}\\
\hspace*{3em}LLM-aware paths select prompt template\dotfill\pageref{LLMselect}\\
\hspace*{3em}Path refinement prompt template\dotfill\pageref{prompt:path_refine}\\
\hspace*{3em}CoT answering evaluation prompt template\dotfill\pageref{prompt:corevaluation}\\
\hspace*{3em}CoT answering generation prompt template\dotfill\pageref{prompt:cot_gen}\\


\twocolumn
\section{Algorithm}
\label{appendix:algorithm}

\subsection{Exploration}\label{appendix:sec:explora}
We summarize the comprehensive algorithmic procedure for evidence exploration detailed in Section~\ref{sec:medthod:exploration} as presented in Algorithm~\ref{algorithm:StructuredRetrieval}-\ref{algorithm:EvidenceExploration}.

\begin{algorithm}
	{
		{
			\SetVline
			\small
			\caption{{{Structured\_Retrieval}}}\label{algorithm:StructuredRetrieval}
            
        \Input{Source KG ($\mathcal{G}$), Question evidence subgraph $\mathcal{G}_q$, select source ($S_c$), topic entities ($\text{List}_{T}$), skyline indicator ($I$), depth ($D$), width ($W$)}
			\Output{Reasoning KG paths ($\text{Paths}_{\text{KG}}$), evidence subgraph($\mathcal{G}_q$)}

           \If {$\text{KG} \in S_c$} {
               
                \If{$\mathcal{G}_q$ is $\emptyset$}
                {
                \State{$\mathcal{G}_q \leftarrow$ Subgraph\_Detection($\mathcal{G}, \text{List}_{T}, D_{\text{max}}$)}
                \State{$\mathcal{G}_q \leftarrow$ KG\_Summary($\mathcal{G}_q$)}
                }
                \State{$ \text{Paths}_{\text{KG}} \leftarrow \text{Tree\_based\_Path\_Retrieval}(Q, I, W, D, \text{List}_{T} ,\mathcal{G}_q)$}
                } 
            \State{\textbf{Return} $\text{Paths}_{\text{KG}}$, $\mathcal{G}_q$}

            \vspace{2mm}
            {\textbf{Tree\_based\_Path\_Retrieval}$(Q, I, W, D_{\max}, \text{List}_{T},\mathcal{G}_q)$\\
            \SetVline
            \small
            \State{$D \leftarrow 1$; $Paths \leftarrow \emptyset$; $E_{\text{outter}} \leftarrow \text{List}_{T}$}
            \While{$D \leq D_{\max}$} {

            \State{$E_{\text{outter}'} \leftarrow \emptyset$}
            \ForEach{$e\in E_{\text{outter}} $} {
                
                \State{$\text{P, outter} \leftarrow  \text{Expand\_One\_Hop}(e)$}
                \State{$\text{Paths} \leftarrow \text{Paths}$ $\cup$ P}
                \State{$E_{\text{outter}'} \leftarrow E_{\text{outter}'}\cup \text{outter}$}
            }
            \While{$|\text{Paths}| > W$} {
                \State{$ \text{Relevant\_Pruning}(\text{Paths},Q, I,W)$}
            
                \State{$E_{\text{outter}'} \leftarrow \text{IntersectMatchUpdate}(\text{Paths}, E_{\text{outter}'} )$}

            }
            \State{$E_{\text{outter}} \leftarrow E_{\text{outter}'}$; $D\leftarrow D+1$}
        }
        \State{\textbf{Return} $\text{Paths}$}
    }
    }
    }
\end{algorithm}
\vspace{-8mm}
\begin{algorithm}
	{
		{
			\SetVline
			\small
\caption{{{Unstructured\_Retrieval}}}\label{algorithm:UnstructuredRetrieval}
            
        \Input{Select source ($S_c$), topic entities ($\text{Topic}(q)$), question ($q$), skyline indicator ($I$), width ($W$)}
        \Output{Summarized wiki structured paths ($\text{Paths}_{\text{Wiki}}$), summarized web  structured paths($\text{Paths}_{\text{Web}}$)}

            \If {$\text{Web} \in S_c$} {
                \State{WebLinks $\leftarrow$ OnlineSearch$(q)$}
                \State{TopURLs $\leftarrow$ $\text{Prompt}_{\text{select}}$(WebLinks, $W, q, I$)}
                \State{Docs $\leftarrow$ URLs\_Process(TopURLs)}
                \State{SelectSentence $\leftarrow$ DRM(Docs, $I$,$W$) }
                \State{Paths$_{\text{Web}}$ $\leftarrow$$\text{Prompt}_{\text{StructuredPathGen}}$(SelectSentence, Topic$(q), I$) }
                }
            \If {$\text{Wiki} \in S_c$} {
                \State{$\textbf{for each}$ e $\in \text{Topic}(q) \textbf{ do }\text{Docs}\leftarrow\text{Docs} \cup \text{Doc}(e)$}
                \State{SelectSentence $\leftarrow$ DRM(Docs, $I$,$W$) }
                \State{Paths$_{\text{Wiki}}$$\leftarrow$$\text{Prompt}_{\text{StructuredPathGen}}$(SelectSentence, Topic$(q), I$) }
                }
            \State{$\text{Prompt}_{\text{PathSummary}}\text{(Paths$_{\text{Wiki}}$, Paths$_{\text{Web}}, I$)}$}
            
            \State{\textbf{Return} Paths$_{\text{Wiki}}$, Paths$_{\text{Web}}$}
            

                
            

    }
    }
\end{algorithm}
\begin{algorithm}[p]
	{
		{
			\SetVline
			\small
			\caption{{{Evidence\_Exploration}}}\label{algorithm:EvidenceExploration}
            
			\Input{Source KG ($\mathcal{G}$),question and split question ($Q = q$ + $q_{\text{split}}$), agentic select source ($S_a$), total available source ($S_t$), topic entities ($\text{Topic}(q)$), skyline indicator ($I_{\text{Sky}}$), predict depth ($D_{\text{predict}}$), maximum depth ($D_{\max}$),  maximum width ($W_{\max}$)}
			\Output{\srag   answers ($a(q)$), final reasoning path ($\text{Paths}_F(q)$)}
            \vspace{3mm}
            
            \CmtState{\\$\text{List}_{T} \leftarrow$ \text{Reorder}($\text{Topic}(q), I_{\text{Sky}}$);\\$D_\text{predict} \leftarrow$ min($D_{\text{predict}}, D_{\max}$); $\mathcal{G}_q \leftarrow \emptyset$} 
            {\textbf{Initial exploration procedure}}
            
               

            \State{$\text{Paths}_{\text{KG}},\mathcal{G}_q  $$\leftarrow \textbf{Structured\_Retrieval}$ $(\mathcal{G},\mathcal{G}_q,S_a, \text{List}_{T},I_{\text{Sky}},D_{\text{predict}},W_{1}$)}
            
            \State{$\text{Paths}_{\text{Wiki}}, \text{Paths}_{\text{Web}} \leftarrow\textbf{Unstructured\_Retrieval}$ $(S_a, \text{Topic}(q)$, $ q, I_{\text{Sky}}, W_{\max}$)}
            \State{$\text{Paths}_{I}\leftarrow$ $\text{Paths}_{\text{KG}}+ \text{Paths}_{\text{Wiki}}+ \text{Paths}_{\text{Web}} $}

            \State{$\text{Paths}_{I} \leftarrow \text{Evidence\_Pruning}(\text{Paths}_{I}, Q, I_{\text{Sky}}, W_{\max},\text{Topic}(q), \mathcal{G}_q)$}
            
            \State{$\text{Answer}, \text{Paths}_{I}\leftarrow \text{Question\_Answering}(\text{Paths}_\text{I}, Q, I_{\text{Sky}})$}

            \State{\textbf{if}  $\texttt{"\{Yes\}"} \text{ in } \text{Answer}$  \textbf{then} \textbf{return} $\text{Answer}, \text{Paths}_{I}$}


\vspace{3mm}
\CmtState{\\$q_{\text{new}}, I_{new} \leftarrow \text{Prompt}_{\text{newQ}}(\text{Paths}_{I}, Q, I_{\text{Sky}}, \text{Topic}(q))$}{\textbf{Refined exploration procedure} }

\State{$\text{Paths}_{\text{KG}},\mathcal{G}_q  $$\leftarrow \textbf{Structured\_Retrieval}$ $(\mathcal{G},\mathcal{G}_q,S_t, \text{List}_{T},I_{\text{new}},D_{\max},W_{1}$)}

\State{$\text{Paths}_{\text{Wiki}}, \text{Paths}_{\text{Web}} \leftarrow\textbf{Unstructured\_Retrieval}$ $(S_t, \text{Topic}(q_{\text{new}})$, $ q_{\text{new}}, I_{\text{new}}, W_{\max}$)}
\State{$\text{Paths}_{R}\leftarrow$ $\text{Paths}_{\text{KG}}+ \text{Paths}_{\text{Wiki}}+ \text{Paths}_{\text{Web}} $}

\State{$\text{Paths}_{R}\leftarrow \text{Evidence\_Pruning}(\text{Paths}_{R}, Q, I_{\text{Sky}}, W_{\max},\text{Topic}(q), \mathcal{G}_q)$}

\State{$\text{Answer}, \text{Paths}_{R}\leftarrow \text{Question\_Answering}(\text{Paths}_\text{R}, Q, I_{\text{Sky}})$}

   \State{\textbf{if}  $\texttt{"\{Yes\}"} \text{ in } \text{Answer}$  \textbf{then} \textbf{return} $\text{Answer}, \text{Paths}_{R}$}

\vspace{3mm}
\CmtState{\\$\text{Paths}_P \leftarrow \emptyset$}{\textbf{Predicted exploration procedure} }

\State{$ \text{Predict}(q) \leftarrow  $LLMPredict($\text{Paths}_{I}+\text{Paths}_{R}, Q, I_{\text{Sky}}$)} 
\ForEach{ $e,I_{\text{Pred}(e)} \in \text{Predict}(q)$} {
    
    \State{$\text{List}_{P} \leftarrow $ Reorder ($\text{Topic}_{q} +e, I_{\text{Pred}(e)}$)}

\State{$\text{Paths}_{\text{KG}},\mathcal{G}_q  $$\leftarrow \textbf{Structured\_Retrieval}$ $(\mathcal{G},\mathcal{G}_q,S_t, \text{List}_{P},I_{\text{Pred}(e)},D_{\max},W_{1}$)}

\State{$\text{Paths}_{\text{Wiki}}, \text{Paths}_{\text{Web}} \leftarrow\textbf{Unstructured\_Retrieval}$ $(S_t, \text{List}_{P}$, $ q, I_{\text{Pred}(e)}, W_{\max}$)}
\State{$\text{Paths}_{P}\leftarrow$ $\text{Paths}_{P}+\text{Paths}_{\text{KG}}+ \text{Paths}_{\text{Wiki}}+ \text{Paths}_{\text{Web}}$}
}

\State{$ \text{Paths}_{P} \leftarrow \text{Evidence\_Pruning}(\text{Paths}_{P}, Q, I_{\text{Sky}}, W_{\max},\text{Topic}(q),\mathcal{G}_q)$}

\State{$\text{Answer}, \text{Paths}_{P}\leftarrow \text{Question\_Answering}(\text{Paths}_\text{P}, Q, I_{\text{Sky}})$}

   \State{\textbf{if}  $\texttt{"\{Yes\}"} \text{ in } \text{Answer}$  \textbf{then} \textbf{return} $\text{Answer}, \text{Paths}_{P}$}

\vspace{3mm}
    \State{
    $\text{Paths}_{F} \leftarrow \text{Paths}_{I}+\text{Paths}_{R}+\text{Paths}_{P}$
    }

\State{$\text{Paths}_{F} \leftarrow \text{Evidence\_Pruning}(\text{Paths}_{F}, Q, I_{\text{Sky}}, W_{\max},\text{Topic}(q), \mathcal{G}_q)$}
    
\State{$\text{Answer}, \text{Paths}_{F}\leftarrow \text{Question\_Answering}(\text{Paths}_\text{F}, Q, I_{\text{Sky}})$}
\State{\textbf{Return} $\text{Answer}, \text{Paths}_{F}$}

}
}
\end{algorithm}

\newpage
\subsection{Evidence pruning}\label{appendix:sec:pruning}
We summarize the comprehensive algorithmic procedure of evidence pruning detailed in Section~\ref{sec:method:prun} as presented in Algorithm~\ref{algorithm:EvidencePruning}. 

\begin{algorithm}
	{
		{
			\SetVline
			\small
			\caption{{{Evidence\_Pruning}}}\label{algorithm:EvidencePruning}
            
			\Input{candidate paths ($C$), question and split question ($Q = q$ + $q_{\text{split}}$), skyline indicator ($I$), width($W_{\max}$), topic entities ($\text{Topic}(q)$), KG ($\mathcal{G}_q$)}
            
			\Output{Pruned candidate paths ($Paths_{c}$)}
            \vspace{3mm}
            \CmtState{\\$S_{\text{rel}},S_{\text{ver}},\text{Cross\_Score}\leftarrow\emptyset$}{\textbf{Step 1: Compute relevance scores}}
            \ForEach{$p_i\in C$}{
            \State{semantic\_score $\leftarrow$ Semantic\_DRM($I$, $p_i$)}
            \State{entity\_overlap $\leftarrow$ Jaccard(Topic($q$), Ent$(p_i)$)}
            \State{$S_{\text{rel}[p_i]}\leftarrow$ $\lambda_{\text{sem}}\cdot \text{semantic\_score} + \lambda_{\text{ent}}\cdot$entity\_overlap}
            }
            \State{$C_{\text{tilde}}$ = Select\_Top\_Paths($C, S_{\text{rel}}, W_1$)}
            \vspace{3mm}
            
            \tcc*[f]{\textbf{Step 2: Compute cross-source verification scores}}
            
            \ForEach{$p_i\in C_{\text{tilde}}$}{
            \State{source\_prior $\leftarrow$ get\_source\_prior($p_i$)}
            \State{supporting\_sources$\leftarrow$ get\_supporting\_sources$(p_i,  C_{\text{tilde}})$}
            \State{source\_agreement$\leftarrow$ $min(|\text{supporting\_sources}|, W_{\max}) / W_{\max}$}
            \State{entity\_alignment $\leftarrow$ |Ent($p_i) \cap E_q|$ / |Ent($p_i)$|}
            \State{$S_{\text{ver}}[p_i$] $\leftarrow \alpha_1\cdot\text{source\_prior}$ + $\alpha_2\cdot \text{source\_agreement}$ + $\alpha_3 \cdot \text{entity\_alignment}$}
            }
            \ForEach{$p_i\in C_{\text{tilde}}$}{
            \State{Cross\_Score[$p_i] \leftarrow $ $\alpha_{\text{cross}} \cdot S_{\text{rel}}[p_i] + (1 - \alpha_{\text{cross}}) \cdot S_{\text{ver}}[p_i]$}
            }

            \State{$\text{Paths}_F$ = Select\_Top\_Paths($C_{\text{tilde}}, \text{Cross\_Score}, W_2$)}
            \vspace{3mm}
            \tcc*[f]{\textbf{Step 3: LLM-aware final selection}}
            
            \State{$\text{Paths}_F$  = Prompt$_{\text{SelectPath}}$($\text{Paths}_F$ ,Q, I, $W_{\max}$)}
            \State{\textbf{Return} $\text{Paths}_F$}

    }
    }
\end{algorithm}

\newpage
~
\newpage

\section{Experiment}
\label{appendix:Additional_expriment}

\subsection{Addtioanl Ablation Study}
\label{appendix:Additional_Ablation}

\begin{figure}[H]
    \begin{subfigure}{0.5\linewidth}
           \centering
           \includegraphics[width=1\linewidth]{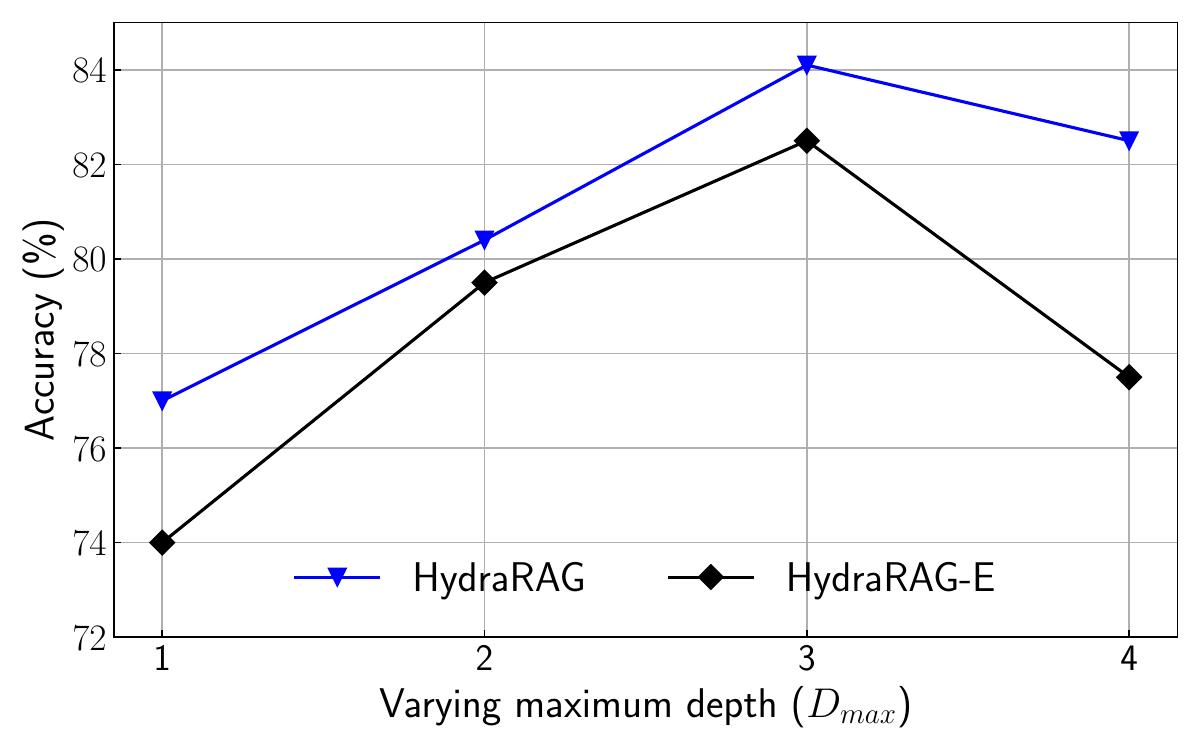}
           \caption{CWQ (Vary $D_{\max}$)}
            \label{fig:a}
    \end{subfigure}
    \begin{subfigure}{0.49\linewidth}
           \centering
           \includegraphics[width=1\linewidth]{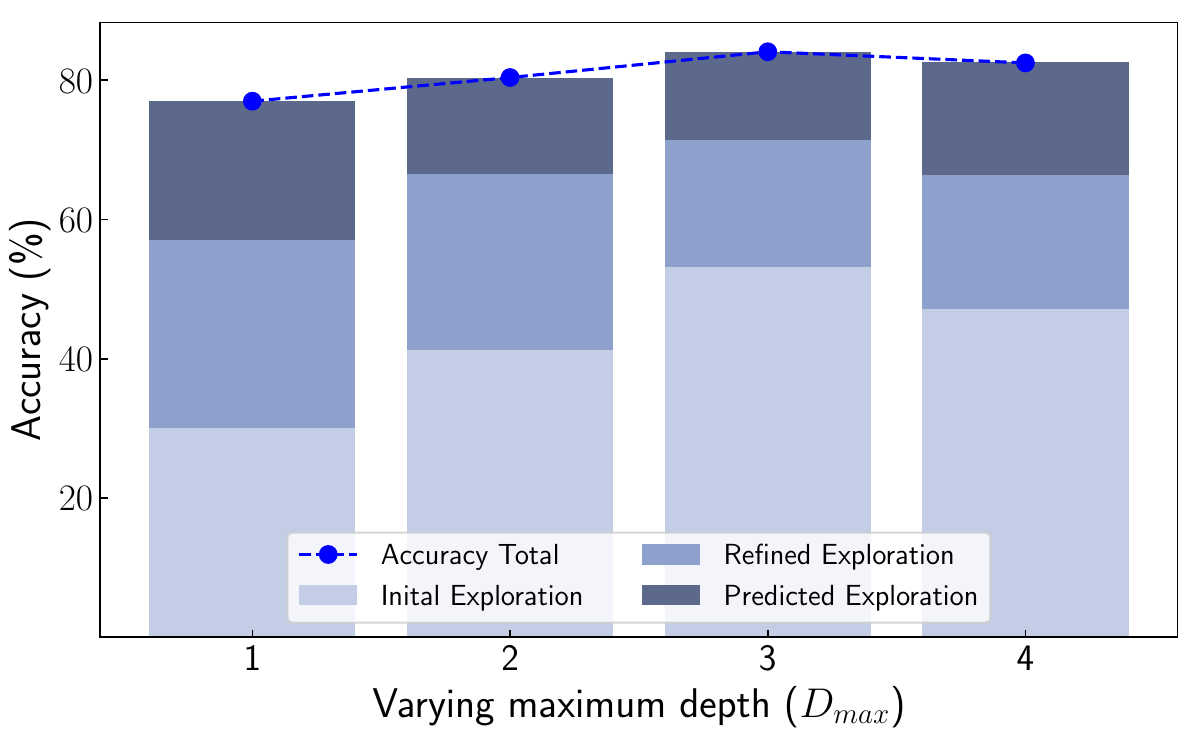}
           \caption{CWQ(\srag  )}
            \label{fig:b}
    \end{subfigure}
    \begin{subfigure}{0.5\linewidth}
           \centering
           \includegraphics[width=1\linewidth]{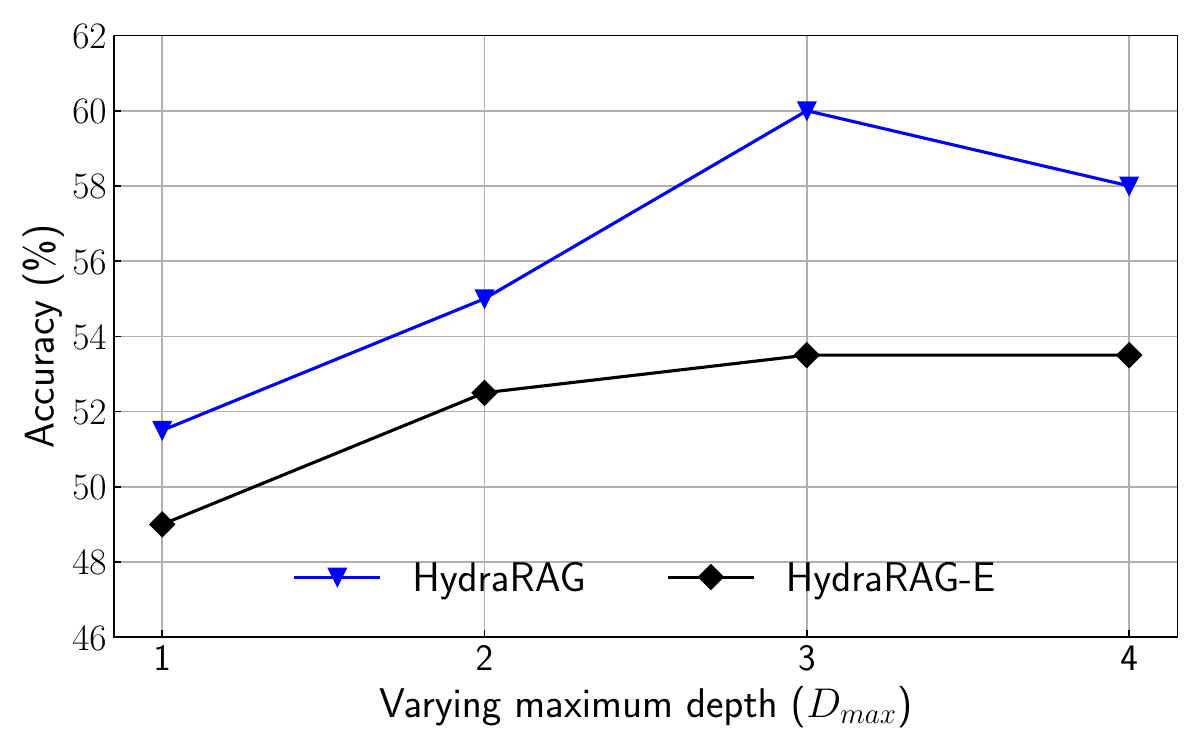}
           \caption{AdvHotpot(Vary $D_{\max}$)}
            \label{fig:c}
    \end{subfigure}
    \begin{subfigure}{0.49\linewidth}
           \centering
           \includegraphics[width=1\linewidth]{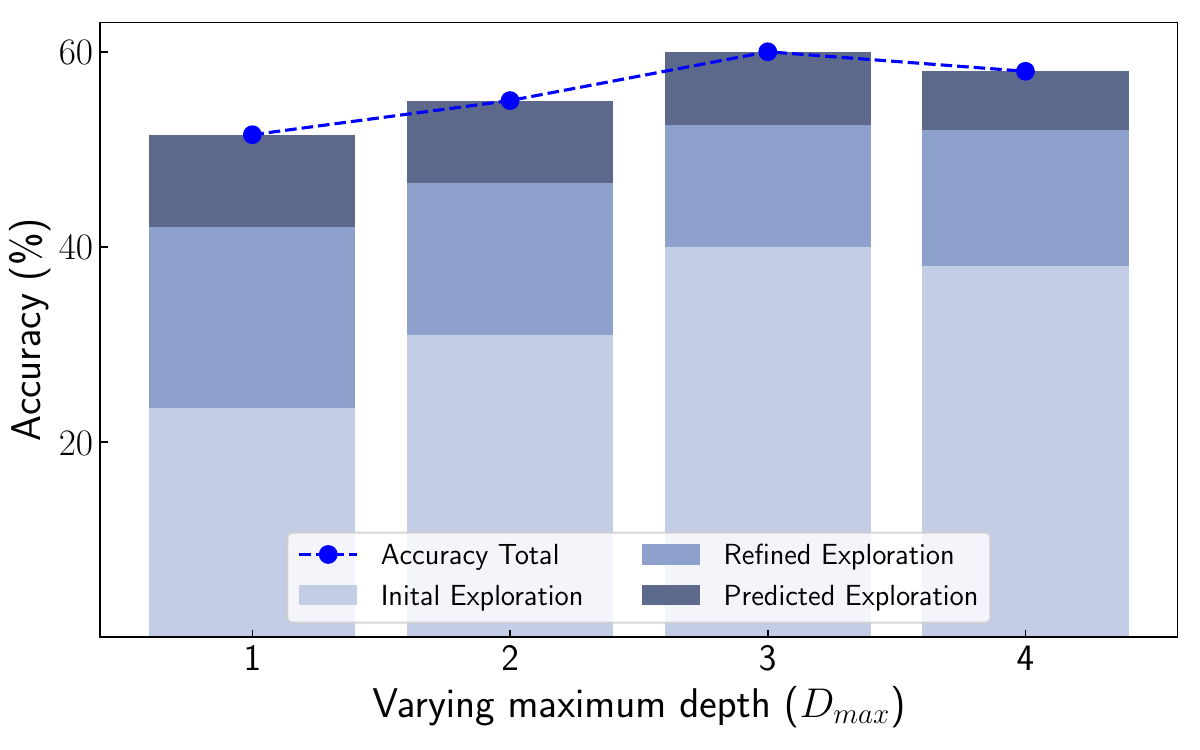}
           \caption{AdvHotpot(\srag  )}
            \label{fig:d}
    \end{subfigure}
\vspace{-2mm}
\caption{The accuracy of \srag   and \srag  -E among CWQ and AdvHotpotQA datasets by varying different $D_{\max}$.}\label{fig:ablation:depth}
\vspace{-1mm}
\end{figure}

\myparagraphquestion{Does search depth matter}  
\label{exp:search_depth}
As described, the dynamic deep search in \srag   is limited by the maximum depth, $D_{\max}$. To analyze how $D_{\max}$ affects performance, we conducted experiments varying depth from 1 to 4. Results (Figures~\ref{fig:ablation:depth}(a) and (c)) show that deeper searches improve performance, but gains diminish beyond depth 3, as excessive depth increases hallucinations and complicates path management. Figures~\ref{fig:ablation:depth}(b) and (d), showing which exploration phase the answer is generated from, reveal that higher depths reduce the effectiveness of both refined and predicted exploration.  Hence, we set $D_{\max}=3$ for optimal balance between performance and efficiency. Notably, even at lower depths, \srag   maintains strong performance by effectively integrating diverse sources and leveraging LLMs' inherent knowledge through the refined and predictive exploration procedures.

\newpage
\myparagraphquestion{How does the agentic source selector affect performance}
\label{exp:agentic_source}
To reduce redundant computational cost when using multiple sources, we incorporate an agentic source selector that adaptively selects sources based on the evolving needs of the reasoning process. To assess its impact, we perform an ablation study comparing \srag and \srag-E with and without the source selector. We evaluate both the accuracy and the average token input during the path pruning stage.
As shown in Table~\ref{fig:ablation:summ_prompt}, integrating the agentic source selector substantially improves performance. For instance, on the CWQ dataset, \srag-E achieves a 24.0\% absolute accuracy improvement, while reducing token input by 36.8\%. Similar trends are observed across other settings.
These improvements stem from the \srag’s ability to dynamically identify and invoke only the most relevant sources. During initial exploration, the selector analyzes the question intent to determine the most suitable sources. In subsequent stages, it further distinguishes between sources used to expand coverage (refined exploration) and those used to increase depth for precise answer prediction (predicted exploration). This adaptive strategy avoids the naïve composition of all sources and leads to more efficient and effective reasoning.
\begin{table}[H]
\caption{Performance comparison of \srag and \srag-E with and without agentic source selector on CWQ and WebQSP datasets.}\label{fig:ablation:summ_prompt}
\resizebox{\linewidth}{!}{
\begin{tabular}{@{}llcc@{}}
\toprule
\textbf{Method}      & \textbf{Evaluation} & \textbf{CWQ} & \textbf{WebQSP} \\ \midrule
\textbf{\srag}         &                     &              &                 \\
w/ agentic source selector   & Accuracy           & \textbf{87.0}        & \textbf{95.0}            \\
                     & Token Input         &   73,240    & 91,097       \\
w/o agentic source selector & Accuracy           & 71.0         & 92.0            \\
                     & Token Input         & 98,748      & 145,411        \\ \midrule
\textbf{\srag-E}       & \textbf{}           & \textbf{}    & \textbf{}       \\
w/ agentic source selector   & Accuracy           & \textbf{85.0 }        & \textbf{92.2}            \\
                     & Token Input         & 73,519      & 97,055         \\
w/o agentic source selector & Accuracy           & 61.0        & 87.0            \\
                     & Token Input         & 116,240      & 49,399     \\ \bottomrule
\end{tabular}
}
\end{table}

\newpage

\newpage
\myparagraphquestion{How do path refinement prompts affect performance}
\label{exp:path refinement}
Inspired by GoT \cite{besta2024graphGoT}, we use path refinement prompts to integrate information from all sources, reduce LLM hallucinations from irrelevant or lengthy paths, and decrease computational costs. To assess their impact, we conduct an ablation study comparing \srag\ and \srag-E with and without path refinement, measuring both accuracy and average token input during path pruning. As shown in Table~\ref{fig:ablation:path_refine}, path refinement increases accuracy by up to 11\% (on CWQ with \srag-E), meanwhile reducing token input by 54\%. These results indicate that path refinement could effectively minimize LLM hallucinations, improve LLM understanding of explored paths, facilitate answer retrieval, enable earlier termination, and reduce overall cost.

\begin{table}[H]
\caption{Performance comparison of \srag and \srag-E with and without path refinement on CWQ and WebQSP datasets.}\label{fig:ablation:path_refine}
\resizebox{\linewidth}{!}{
\begin{tabular}{@{}llcc@{}}
\toprule
\textbf{Method}      & \textbf{Evaluation} & \textbf{CWQ} & \textbf{WebQSP} \\ \midrule
\textbf{\srag}         &                     &              &                 \\
w/ Path refinement   & Accuracy           & \textbf{87.0}        & \textbf{95.0}            \\
                     & Token Input         &   73,240    & 91,097       \\
w/o Path refinement & Accuracy           & 79.0         & 93.0            \\
                     & Token Input         & 134,554      & 107,516         \\ \midrule
\textbf{\srag-E}       & \textbf{}           & \textbf{}    & \textbf{}       \\
w/ Path refinement   & Accuracy           & \textbf{85.0 }        & \textbf{92.2}            \\
                     & Token Input         & 73,519      & 97,055         \\
w/o Path refinement & Accuracy           & 74.0        & 90.0            \\
                     & Token Input         & 159,678      & 107,762         \\ \bottomrule
\end{tabular}
}
\end{table}

\begin{table*}
\centering
\caption{Performance comparison of \srag with different knowledge sources and retrieval components across four multi-hop datasets.
}
\label{tab:ablation_sources}
\resizebox{0.7\linewidth}{!}{
\begin{tabular}{lcccc}
\toprule
\textbf{Source Setting} & \textbf{CWQ} & \textbf{AdvHotpotQA} & \textbf{WebQSP} & \textbf{QALD} \\
\midrule
w/ all sources & 78.2 & 55.2 & 90.3 & 78.0 \\
w/o Freebase & 63.7 & 51.0 & 79.2 & 70.0 \\
w/o WikiKG & 70.0 & 54.0 & 86.7 & 69.0 \\
w/o Web document& 75.0 & 52.3 & 86.0 & 75.3 \\
w/o Wiki document & 74.0 & 50.4 & 84.0 & 68.2 \\
w/o Freebase \& WikiKG  & 60.4 & 50.1 & 74.0 & 64.0 \\
w/o Web \& Wiki document & 73.6 & 42.4 & 86.0 & 71.7 \\
\bottomrule
\end{tabular}
}
\end{table*}
\newpage

\myparagraphquestion{How do different knowledge sources affect the performance of \srag}
\label{exp:diifferent knowledge sources affect the perfor}
To evaluate the impact of different knowledge sources and retrieval components, we conduct ablation experiments by excluding individual sources and modules on all multi-hop QA datasets. Results show that Freebase contributes most to CWQ and WebQSP, while WikiKG and wiki documents are more important for AdvHotpotQA and QALD, likely due to varying knowledge backgrounds and overlaps in each dataset. Notably, removing any single source or retrieval module does not cause a dramatic drop in performance, demonstrating the robustness of our framework in integrating heterogeneous evidence. \srag\ effectively leverages complementary information from both structured and unstructured sources, mitigating the impact of missing components. Even without structured retrieval (Freebase and WikiKG), \srag\ maintains high accuracy and still outperforms naive text and web-based RAG methods using the same corpus. This highlights the strength of our structure-aware integration in extracting and organizing information from unstructured evidence, bridging the gap between text-based and structure-based approaches. Overall, these results underline the benefit of our unified multi-source framework, which ensures stable, high performance by flexibly combining evidence from diverse sources.


\newpage

\newpage
\subsection{Additional Effectiveness Evaluation}
\label{appendix:Additional_Effectiveness}

\myparagraph{Effectiveness on cross-source verification}
\label{exp:cross-source_verification}
To evaluate the effectiveness of cross-source verification, we compare it with the standard question-relevant approach commonly used in hybrid RAG. For a fair comparison, we use the same embedding model (SBERT) and beam search width $W_2$, replacing only the first two evidence pruning steps (source relevance and cross-source verification). We report accuracy, average token input, and number of LLM calls for both pruning strategies on the CWQ and AdvHotpotQA datasets (Table~\ref{table:multi-s-evaluation_results}). Results show that 
cross-source verification improves accuracy by up to 22\% on CWQ and reduces token cost by up to 41.8\% on AdvHotpotQA, using the same knowledge corpus. 
 This improvement arises because relevance-only pruning often retains noisy paths and prunes correct ones, forcing extra exploration and incurring higher LLM costs. 
These results demonstrate the effectiveness of cross-source verification and its potential as a solution for efficient multi-source RAG.
\begin{table}[H]
\caption{Evaluation Results for CWQ and AdvHotpotQA with cross-source verification and question relevance pruning.}
\centering
\label{table:multi-s-evaluation_results}
\setlength\tabcolsep{1.0 pt}
\resizebox{1\linewidth}{!}{
\centering
\begin{tabular}{@{}llcc@{}}
\toprule
\textbf{Method}         & \textbf{Evaluation}          & \textbf{CWQ}  & \textbf{AdvHotpotQA} \\ \midrule
w/ Cross-source           & Accuracy                & 84.0          & 60.0            \\
verification                             & Token Input     & 114,023       & 14,089          \\
                            & LLM Calls       & 8.0           & 9.0             \\ \midrule
w/ Question    & Accuracy                & 62.0          & 52.1            \\
Relevant Only                            & Token Input     & 157,850       & 24,193          \\
                            & LLM Calls       & 7.9           & 9.6             \\ \bottomrule
\end{tabular}
}
\end{table}

\newpage

\begin{figure}[H]
    \centering
    \includegraphics[width=0.9\linewidth]{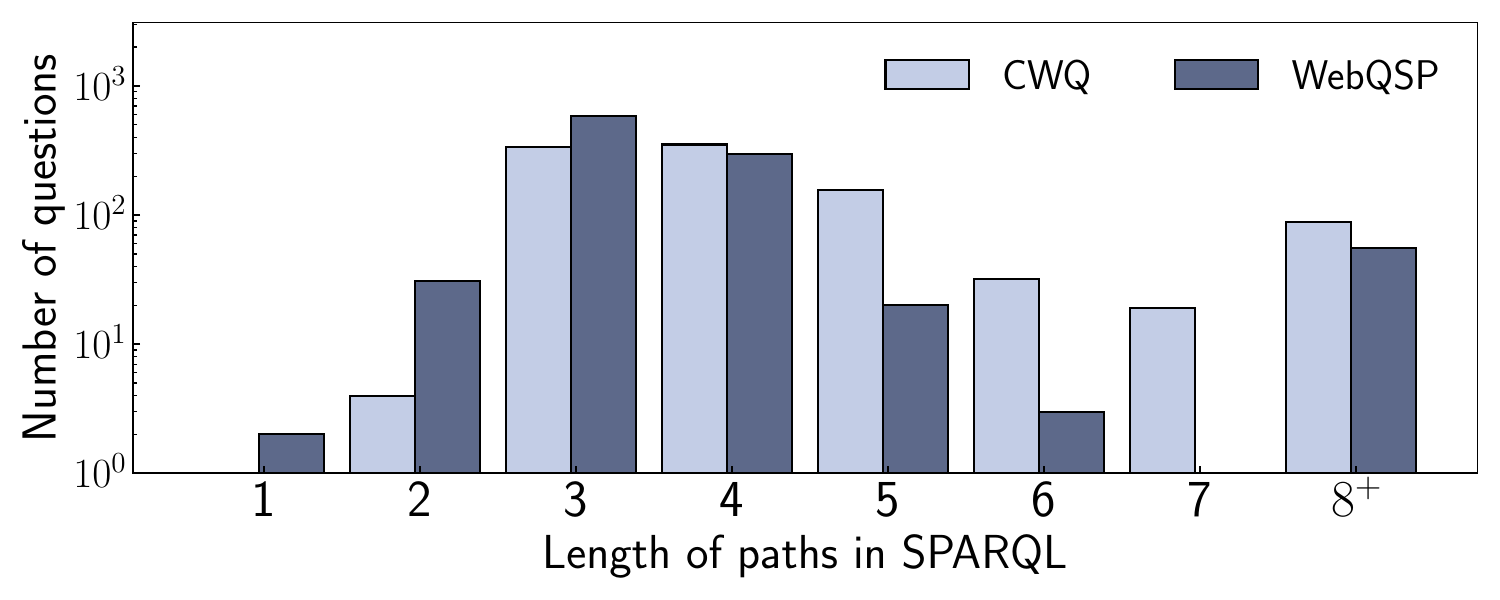}
    \vspace{-2mm}
    \caption{The lengths of the ground-truth SPARQL queries within the CWQ and WebQSP datasets.}
    
    \label{fig:exp:q_by_length}

    \begin{subfigure}{0.48\linewidth}
           \centering
           \includegraphics[width=1\linewidth]{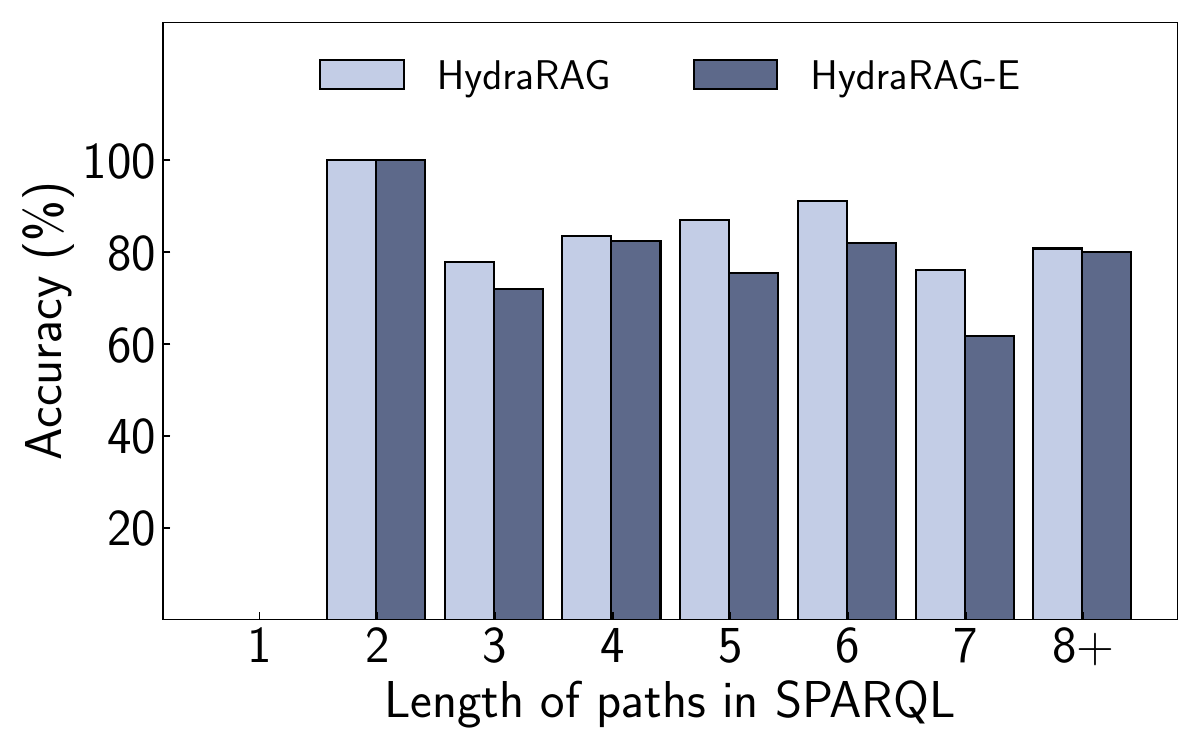}
           \caption{CWQ}
            \label{fig:aa}
    \end{subfigure}
    \begin{subfigure}{0.48\linewidth}
           \centering
           \includegraphics[width=1\linewidth]{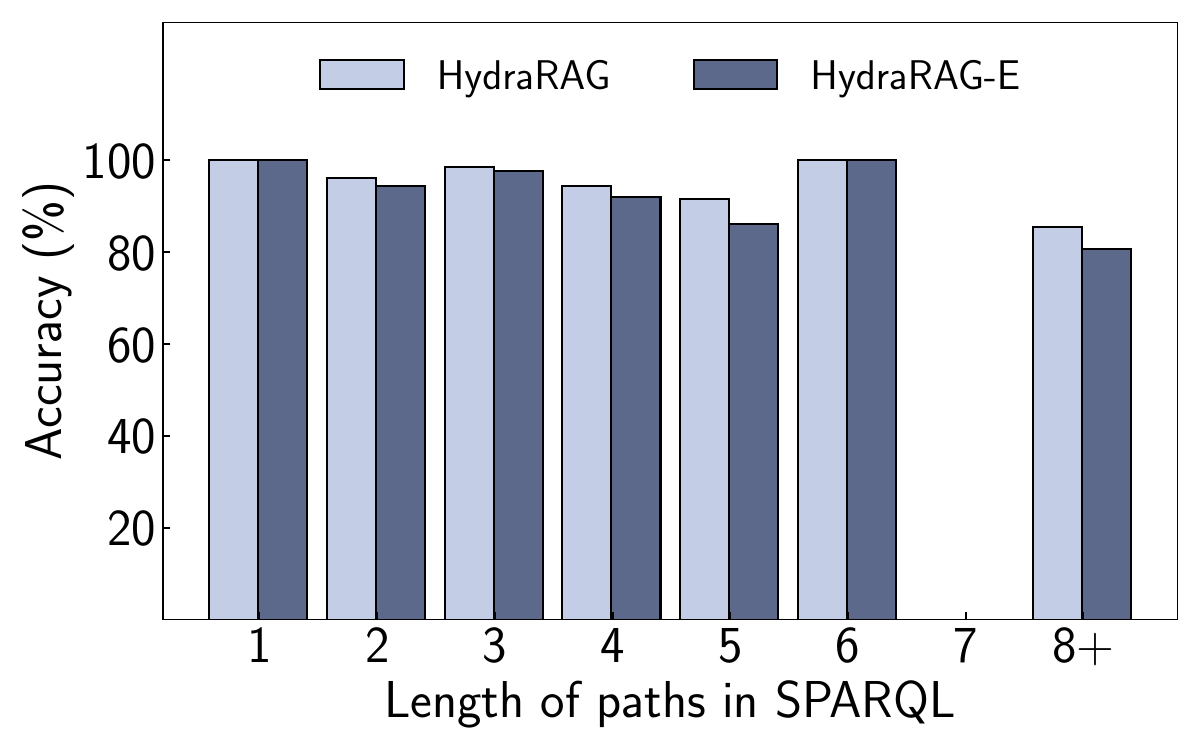}
           \caption{WebQSP}
            \label{fig:ab}
    \end{subfigure}
    \vspace{-2mm}
\caption{Accuracy of \srag and \srag-E on the CWQ and WebQSP datasets, categorized by the different lengths of the ground-truth answers for each question.}\label{fig:exp:query_by_length}
\end{figure}

\myparagraph{Effectiveness on multi-hop reasoning}
\label{exp:multi-hop}
To assess \srag’s performance on multi-hop reasoning tasks, we analyze accuracy by grouping questions according to the length of their ground-truth SPARQL queries. We randomly sample 1,000 questions each from the CWQ and WebQSP datasets and determine reasoning length by counting the number of relations in each ground-truth query (see Figure~\ref{fig:exp:q_by_length}). We then evaluate \srag and \srag-E across varying reasoning lengths to understand their effectiveness under varying query complexities. As shown in Figure~\ref{fig:exp:query_by_length}, both models maintain high and stable accuracy across different lengths, with \srag achieving up to 98.6\% accuracy even at the highest length levels in WebQSP. Notably, \srag can correctly answer questions with ground-truth lengths of eight or more by exploring novel paths and integrating LLM knowledge, rather than strictly matching the ground-truth path. These results highlight the effectiveness of \srag in handling complex multi-hop reasoning tasks.

\begin{table*}
\centering
\caption{Average number of entities from Freebase, WikiKG, and after graph fusion and reduction for three datasets.}
\label{tab:entity_fusion_reduction}
\setlength\tabcolsep{1.0 pt}
\resizebox{0.7\linewidth}{!}{
\begin{tabular}{lccc}
\toprule
 & \textbf{CWQ} & \textbf{AdvHotpotQA} & \textbf{QALD10-en} \\
\midrule
Ave. Entity Number from Freebase & 2,289,881 & 1,329,012 & 2,753,230 \\
Ave. Entity Number from WikiKG& 160,762 & 128,766 & 389,360 \\
Ave. Entity Number after Reduction & 128,352 & 399,785 & 587,110 \\
\bottomrule
\end{tabular}
}
\vspace{-1mm}
\end{table*}

\begin{table*}
\centering
\caption{Performance of \srag and \srag-E on multi-entity and single-entity questions of all datasets. The symbol `-' indicates no multi-entity question inside.}
\label{fig:exp:mult_e}
\setlength\tabcolsep{1.0 pt}
\resizebox{0.7\linewidth}{!}{
\begin{tabular}{@{}lccccccc@{}}
\toprule
\textbf{Question Set} & \textbf{CWQ} & \textbf{WebQSP} & \begin{tabular}[c]{@{}c@{}}\textbf{AdvHotpot}\\\textbf{QA}\end{tabular} & \begin{tabular}[c]{@{}c@{}}\textbf{QALD10}-\\\textbf{en}\end{tabular} & \begin{tabular}[c]{@{}c@{}}\textbf{Simple}\\\textbf{Questions}\end{tabular} & \begin{tabular}[c]{@{}c@{}}\textbf{ZeroShot}\\\textbf{RE}\end{tabular} & \begin{tabular}[c]{@{}c@{}}\textbf{Web}\\\textbf{Questions}\end{tabular}  \\ \midrule
\multicolumn{8}{@{}l@{}}{$\textbf{\srag}$ w/ GPT-3.5-Turbo} \\
Single-entity & 71.9 & 96.2 & 56.8 & 83.1  & 89.0 & 97.7 & 88.2 \\
Multi-entity  & 92.0 & 93.1 & 61.5 & 86.5  & -    & 83.6 & 82.8 \\ \midrule
\multicolumn{8}{@{}l@{}}{\textbf{\srag-E} w/ GPT-3.5-Turbo} \\
Single-entity & 68.6 & 94.0 & 51.5 & 79.1 & 87.3 & 97.3 & 85.4  \\
Multi-entity  & 89.7 & 89.7 & 57.1  & 84.2 & -    & 80.3 & 82.8 \\
\bottomrule
\end{tabular}}
\vspace{-1mm}
\end{table*}
\newpage

\myparagraph{Effectiveness on graph structure pruning}
\label{exp:structure_pruning}
To assess the effectiveness of our graph fusion and reduction strategy, we report the average number of unique entities from Freebase and WikiKG before fusion, and the total number of entities remaining after fusion and graph reduction, as shown in Table~\ref{tab:entity_fusion_reduction}. For each dataset, we first fuse overlapping entities from multiple knowledge sources, then apply the graph reduction method described in Section~\ref{sec:med:initial} to remove irrelevant nodes prior to path exploration.
The results demonstrate a substantial reduction in the number of entities across all datasets. For example, in CWQ, the initial combined entity count from Freebase and WikiKG exceeds 2.4 million, but this is reduced to only 128,352 after fusion and pruning. Similar trends are observed for AdvHotpotQA and QALD10-en. This reduction indicates that a significant portion of entities are either redundant or irrelevant to the questions under consideration. By eliminating such entities before downstream reasoning, our approach improves computational efficiency and focuses exploration on the most relevant subgraphs. Overall, these results verify the effectiveness of combining graph fusion and reduction for constructing compact and informative question-specific subgraphs.
\newpage
\myparagraph{Effectiveness on multi-entity questions}
\label{exp:multi-entity}
Graphs are widely used to model complex relationships among different entities \cite{chen2025covering,chen2024querying,CHenYIn,DBLP:journals/pvldb/WuSWWZQL24,zhang2023hop}.
Recent advances have also explored calibration, label shift estimation, and temporal benchmarks in time series and classification tasks \cite{zhang2025instance,zhang2025label,zhang2025revisit}.
KGs store triples, making entity links explicit \cite{yizhang1,Yizhang2,zhai2024adapting,zhai2025graph,yin2025efficient}.
Building on this foundation, we further examine how well \srag can leverage such structural representations when dealing with questions that involve multiple entities.

To evaluate the performance of \srag on multi-entity questions, we report the accuracy on all test sets by categorizing questions based on the number of topic entities. The results, shown in Table~\ref{fig:exp:mult_e}, demonstrate that, despite the increased complexity of multi-entity questions compared to single-entity ones, \srag maintains excellent accuracy, achieving up to 93.1\% on the WebQSP dataset. This underscores the effectiveness of our structure-based model in handling complex multi-entity queries.

\newpage
\subsection{Reasoning Faithfulness Analysis}
\label{appendix:fathhful_analysis}

\begin{figure}[h]
    \centering
    \includegraphics[width=1\linewidth]{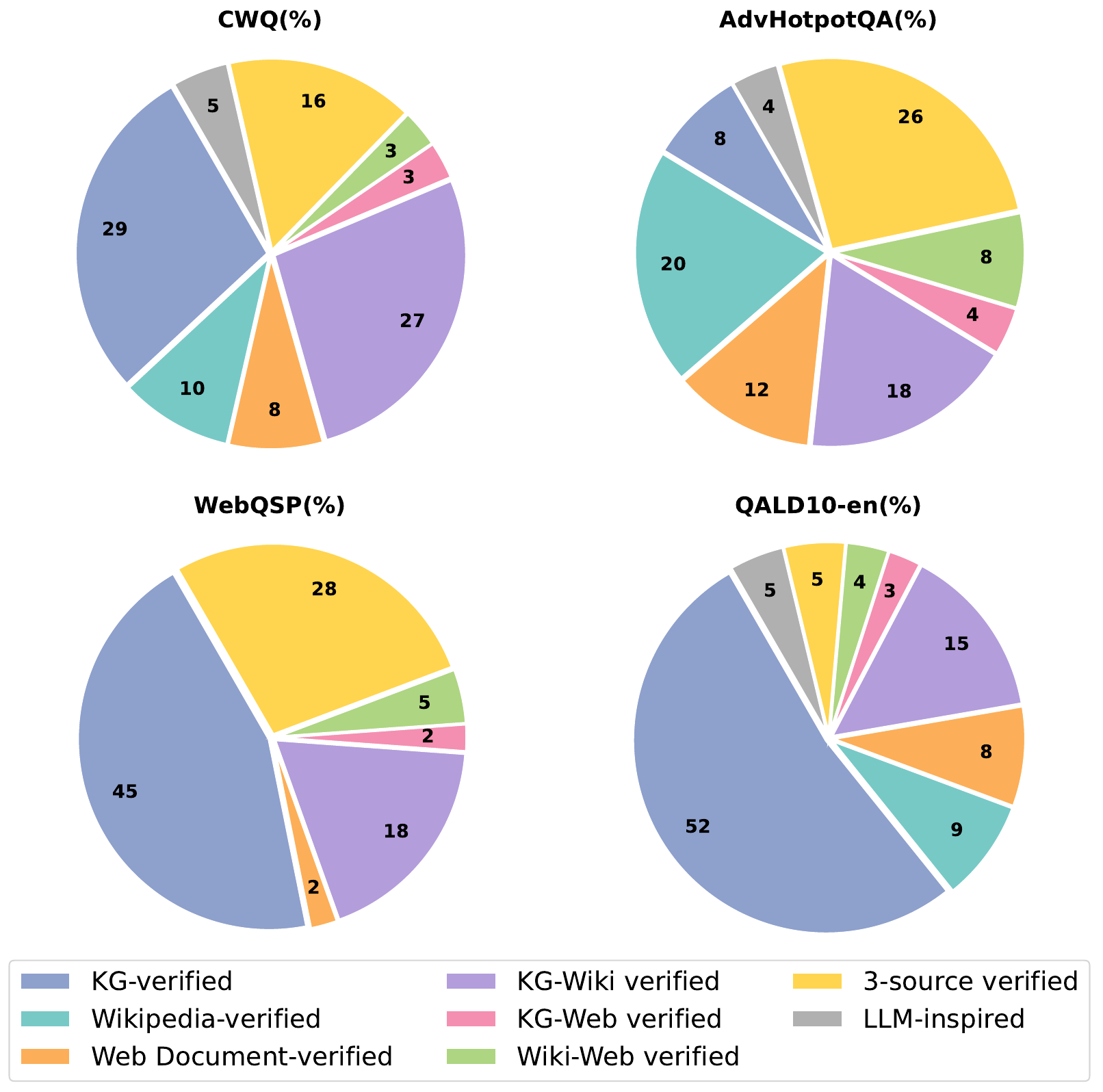}
    \caption{The proportions of answer evidence and cross validation of \srag among CWQ, WebQSP, AdvHotpotQA, and QALD10-en datasets.}\label{fig:exp:answer_gen_by}
\end{figure}

\myparagraph{Evidence of answer exploration sources}
\label{exp:Exploration Source}
We analyze the sources of evidence supporting correct answers on four multi-hop datasets to assess the effectiveness of cross-verification and the distribution of knowledge supervision in \srag, as shown in Figure~\ref{fig:exp:answer_gen_by}. Specifically, all generated answers are classified based on the verification source: KG-verified, Wikipedia-verified, web document-verified, as well as combinations such as KG-Wiki, KG-Web, Wiki-Web, and those verified by all three sources. In addition, when the paths generated from all external sources are insufficient to reach the answer, and the LLM supplements the reasoning using its inherent knowledge, such answers are categorized as LLM-inspired.
The analysis reveals that over 95\% of answers are supported by external knowledge supervision, confirming that \srag primarily grounds its reasoning in verifiable sources. Furthermore, up to 56\% of correct answers are jointly verified by at least two distinct knowledge sources. 
This highlights the strength of \srag in leveraging multi-source evidence, an essential for faithful and interpretable reasoning.

Among answers with only single-source support, knowledge graph (KG) evidence dominates, accounting for as much as 95.7\% of sole-source supervision in WebQSP. This underscores the high reliability and factual precision of KGs compared to other sources.
Compared with previous methods that simply combine LLM internal knowledge with external sources~\cite{tog2.0ma2024think}, \srag further enhances reliability by enabling mutual cross-verification between all sources. This multi-source evaluation mechanism reduces the risk of unsupported or spurious answers.
These results highlight that \srag is a faithful reasoning framework that not only prioritizes evidence-based answers but also ensures high accuracy and interpretability by integrating and cross-validating structured and unstructured knowledge.
\begin{figure}[h]
    \begin{subfigure}{0.495\linewidth}
           \centering
           \includegraphics[width=1\linewidth]{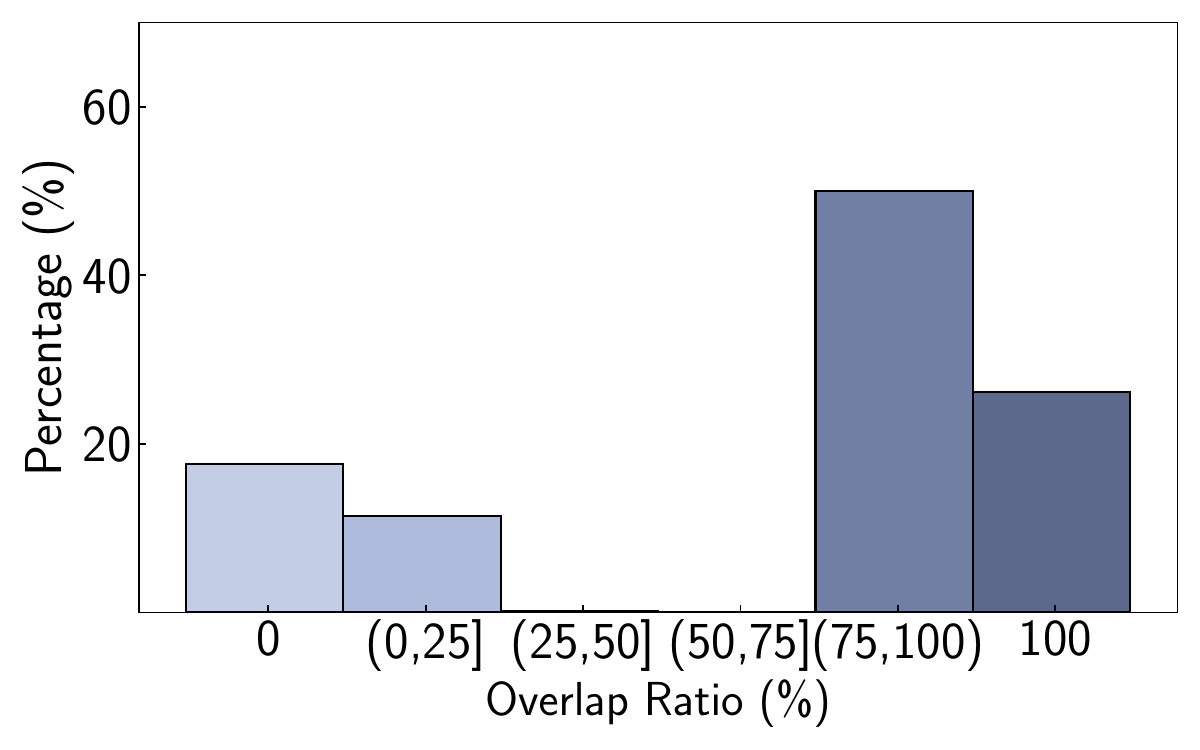}
           
           \caption{CWQ (\srag)}
            \label{fig:a}
    \end{subfigure}
    \begin{subfigure}{0.495\linewidth}
           \centering
           \includegraphics[width=1\linewidth]{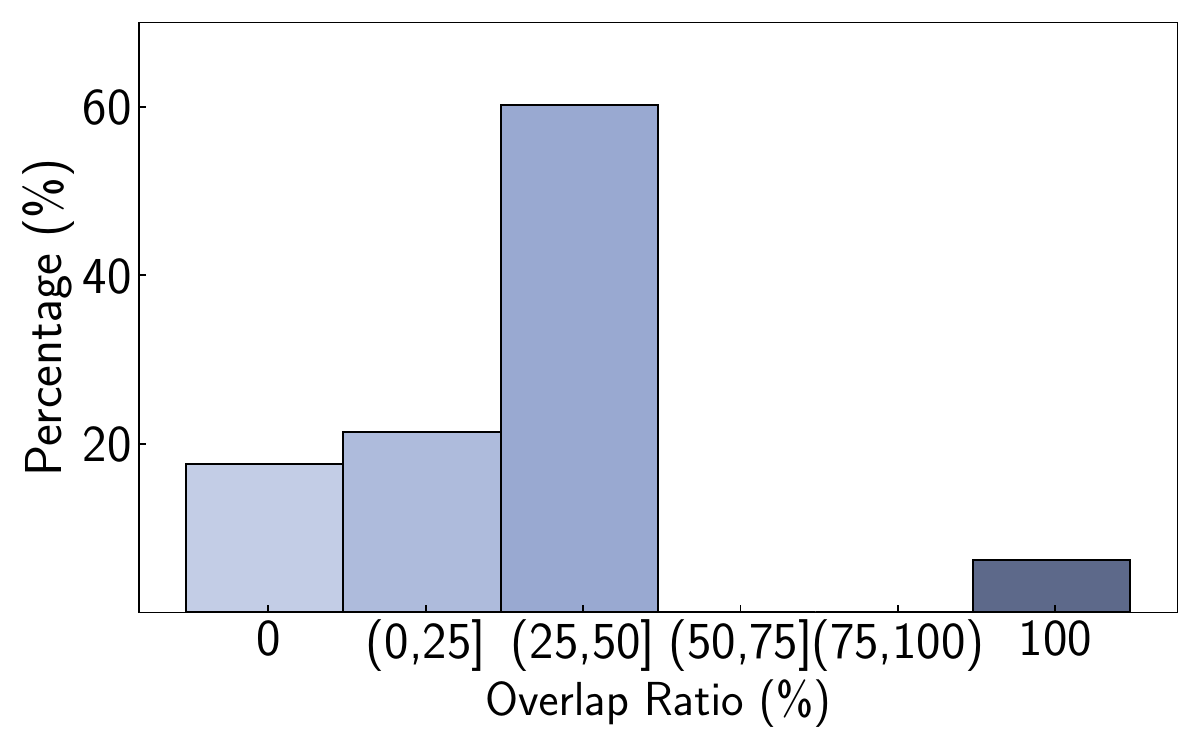}
           
           \caption{CWQ (\srag-E)}
            \label{fig:a}
    \end{subfigure}
    \begin{subfigure}{0.495\linewidth}
           \centering
           \includegraphics[width=1\linewidth]{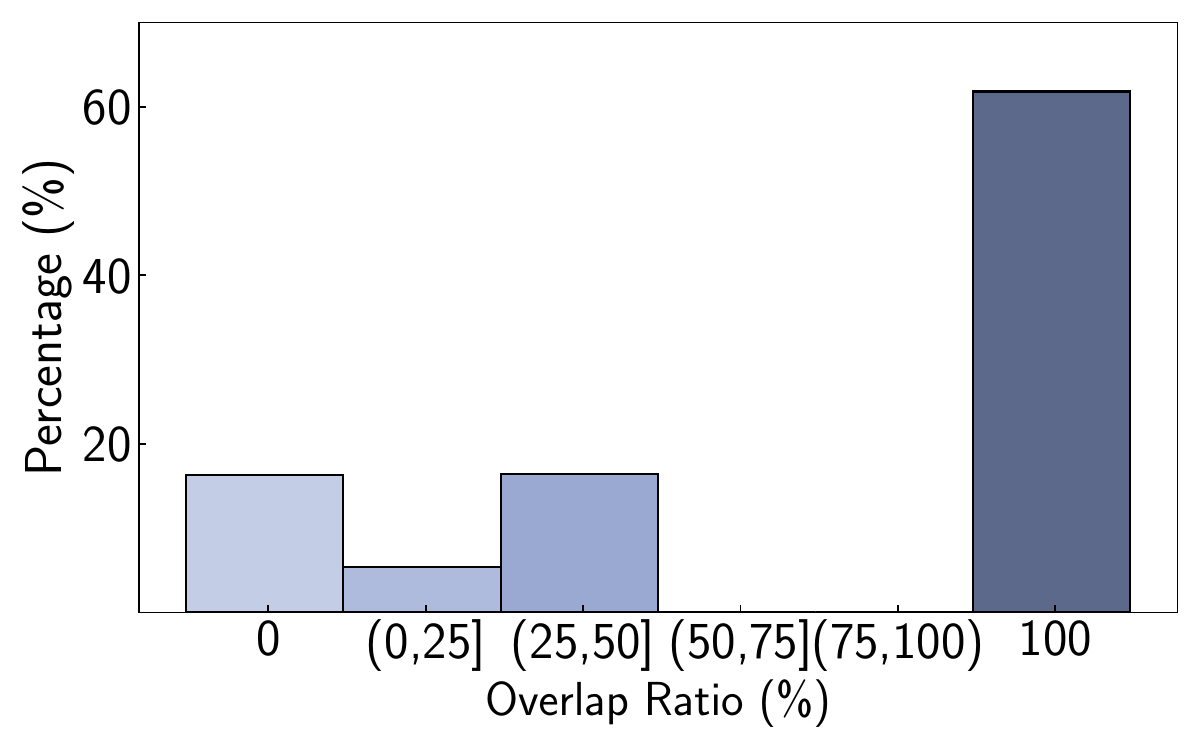}
           
           \caption{WebQSP (\srag)}
            \label{fig:a}
    \end{subfigure}
    \begin{subfigure}{0.495\linewidth}
           \centering
           \includegraphics[width=1\linewidth]{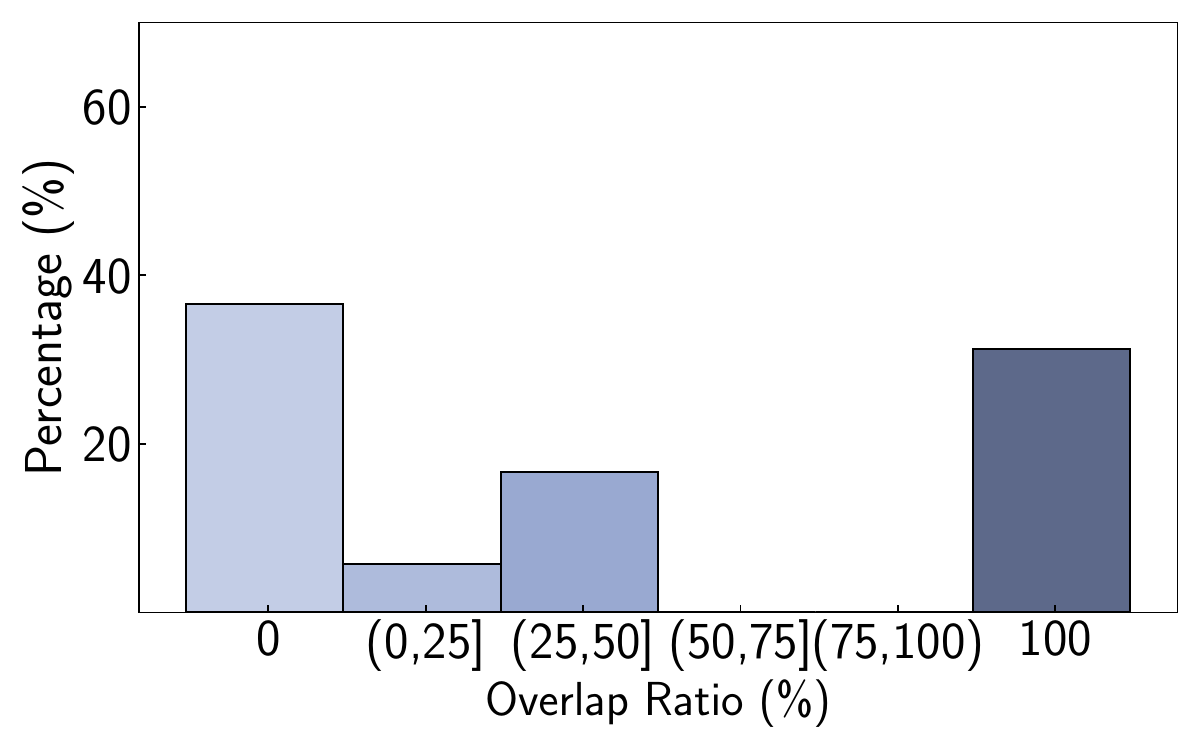}
           
           \caption{WebQSP (\srag-E)}
            \label{fig:a}
    \end{subfigure}

\caption{ The path overlap ratio of \srag and \srag-E among CWQ, and WebQSP datasets.}\label{fig:exp:overlap}
    
\end{figure}

\myparagraph{Overlap ratio between explored paths and ground-truth paths}
\label{exp:Overlapground-truth paths}
We analysis correctly answered samples from CWQ and WebQSP to examine the overlap ratio between paths $P$ explored by \srag and ground-truth paths $P_G$ from SPARQL queries. The overlap ratio is defined as the proportion of shared relations to total relations in the ground-truth SPARQL path:

\[
Ratio(P) = \frac{|Relation(P) \cap Relation(P_G)|}{|Relation(P_G)|},
\]
where $Relation(P)$ is the set of relations in path $P$. Figure~\ref{fig:exp:overlap} shows the distribution of overlap ratios. For WebQSP, \srag achieves the highest proportion of fully overlapping paths (about 61\%), while \srag-E shows the most paths with up to 37\% non-overlapping relations, indicating that \srag-E explores novel paths to derive the answers. This difference is due to \srag-E's approach of randomly selecting one related edge from each cluster. These results highlight the effectiveness of our structure-based exploration in generating both accurate and diverse reasoning paths.


\subsection{Error Analysis}\label{appendix:exp:error_analysis}

To further examine the integration of LLMs with KGs, we conduct an error analysis on the CWQ, WebQSP, and GrailQA datasets. Errors are categorized into four types: (1) answer generation errors, (2) refusal errors, (3) format errors, and (4) other hallucination errors. An answer generation error is defined as the case where \srag provides a correct reasoning path, but the LLM fails to extract the correct answer from it.

\begin{figure}[H]
    \centering
    \includegraphics[width=1\linewidth]{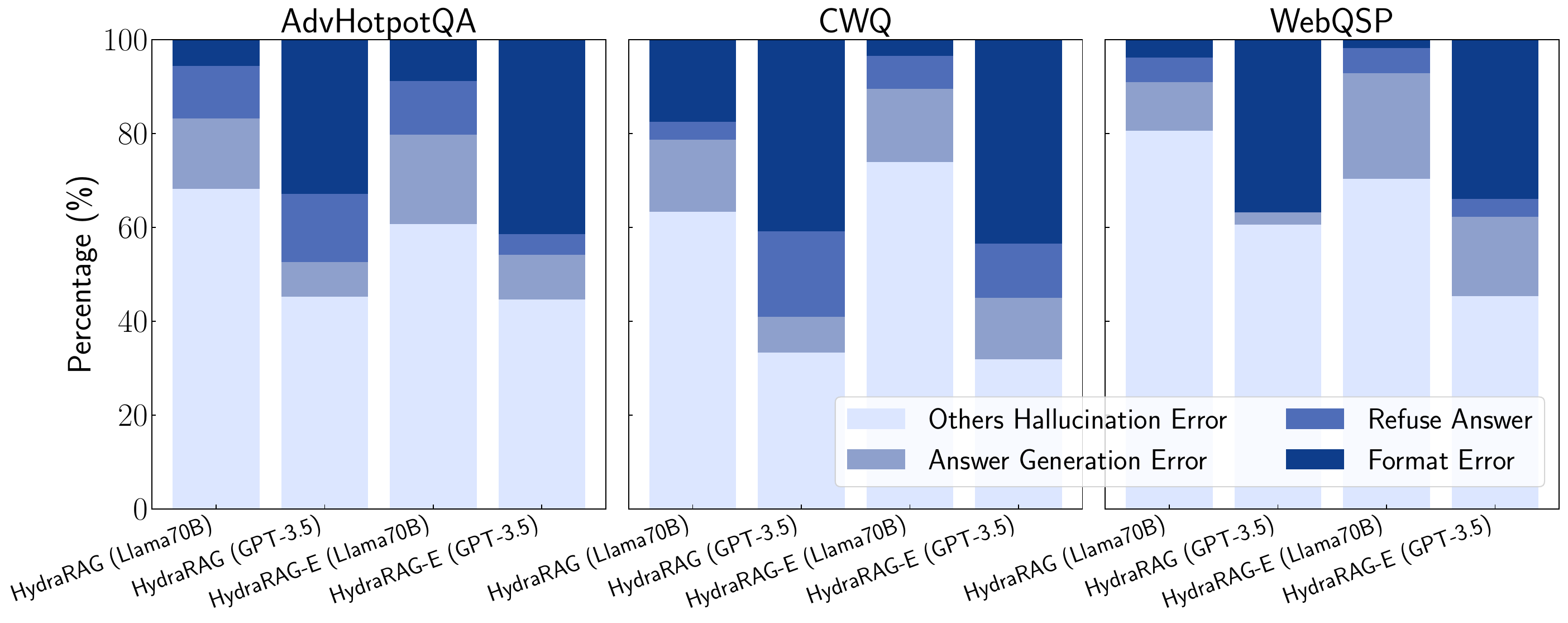}
    \caption{The error instances and categories of \srag and \srag-E in the AdvHotpotQA, CWQ, and WebQSP datasets.}
    \vspace{-3mm}
    \label{fig:exp:error_analysis}
\end{figure}

Figure~\ref{fig:exp:error_analysis} shows the distribution of these error types. The results indicate that more advanced LLMs generally reduce the incidence of "other hallucination errors", "refusal errors", and "answer generation errors", as improved reasoning capabilities allow the model to make better use of the retrieved data. The reduction in "answer generation errors" in particular demonstrates that advanced LLMs can more effectively utilize the reasoning paths generated by \srag. However, we also observe an increase in "format errors" with stronger LLMs, which may be due to their increased creative flexibility in generating outputs.

\subsection{Efficiency Analysis}\label{appendix:effiency_analysis}
\begin{figure}
    \centering
    \includegraphics[width=1\linewidth]{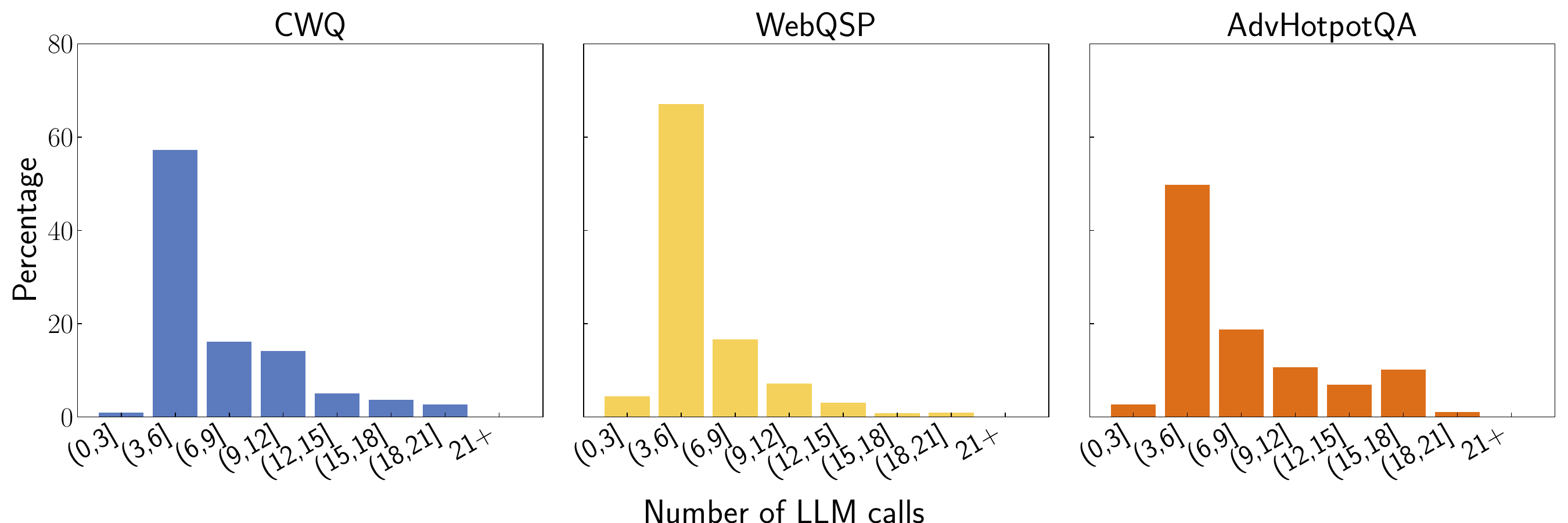}
    
    \caption{The proportion of questions of \srag and \srag-E by different LLM Calls among CWQ, WebQSP, and AdvHotpotQA datasets.}
    \label{fig:LLM_call_percentage}
\end{figure}

\myparagraph{LLM calls cost analysis}
\label{exp:LLM calls cost}
To evaluate the cost and efficiency of utilizing LLMs, we conducted an analysis of LLM calls on the  CWQ,  WebQSP, and AdvHotpotQA datasets. Initially, we examined the proportion of questions answered with varying numbers of LLM calls, as depicted in Figure~\ref{fig:LLM_call_percentage}. The results indicate that the majority of questions are answered within nine LLM calls across all datasets, with approximately 60\% and 70\% of questions being resolved within six calls on CWQ and WebQSP, respectively. These findings demonstrate \srag's efficiency in minimizing LLM costs.

\begin{table}
\centering
\vspace{-1mm}
\caption{Efficiency analysis of different methods on AdvHotpotQA.}
\vspace{-1mm}
\label{tab:efficiency_advhotpot}
\resizebox{1\linewidth}{!}{

\begin{tabular}{lccc}
\toprule
\textbf{Method} & \textbf{Average Total Time} & \textbf{API Calls} & \textbf{Accuracy} \\
\midrule
\srag & 43.0 & 8.7 & 60.7 \\
ToG-2 & 27.3 & 5.4 & 42.9 \\
ToG & 69.3 & 16.3 & 26.3 \\
CoK & 30.1 & 11.0 & 45.4 \\
\bottomrule
\end{tabular}
}
\vspace{-3mm}
\end{table}

\newpage
\myparagraph{Efficiency analysis on AdvHotpotQA}
\label{exp:Efficiency Analysis on AdvHotpotQA}
We compare the efficiency and effectiveness of different multi-hop QA methods on the AdvHotpotQA dataset by reporting average processing time, number of API calls per question, and answer accuracy, as shown in Table~\ref{tab:efficiency_advhotpot}. Among all methods, \srag achieves the highest accuracy (60.71\%) while maintaining a moderate average total processing time (43 seconds) and relatively low API call cost (8.7 per question). Compared to ToG-2 and CoK, which exhibit lower accuracy (42.9\% and 45.4\%, respectively), \srag offers a clear advantage in answer quality without excessive time or API usage. While ToG-2 achieves the lowest average time and API calls, its accuracy lags significantly behind \srag. Conversely, ToG has the highest processing time and API usage with the lowest accuracy among all compared methods.
These results demonstrate that \srag effectively balances efficiency and answer quality, providing a more accurate solution than previous methods while controlling computation and LLM call costs.


\newpage
\section{Experiment Details}
\label{appen:dataset_details}

\myparagraph{Experiment datasets}
\label{exp:Experiment datasets}
To evaluate the capability of \srag on complex, knowledge-intensive reasoning tasks, we evaluate it on seven KBQA benchmarks. These include four multi-hop datasets: ComplexWebQuestions (CWQ) \cite{talmor-berant-2018-web}, WebQSP \cite{yih-etal-2016-value}, AdvHotpotQA \cite{ye2022unreliability}, and QALD10-en \cite{usbeck2024qald}, a single-hop dataset: Simple Questions (SimpleQA) \cite{simplequestions}, a slot filling dataset: ZeroShot RE \cite{petroni2020kilt}, and an open-domain QA dataset: WebQuestions \cite{berant-etal-2013-semantic},  to examine \srag on more general tasks.
For fair comparison with strong prompt-based baselines, we use the same test splits reported in \cite{pogtan2025paths,tog1.0sun2023think,tog2.0ma2024think}. 

As background knowledge, we employ the full Freebase \cite{freebase}, Wikipedia, and Wikidata \cite{vrandevcic2014wikidata}. Using the complete knowledge setting, rather than a distractor subset, makes retrieval more challenging and better evaluates each method’s reasoning ability \cite{tog2.0ma2024think}. The statistics of the datasets utilized in this paper are detailed in Table~\ref{fig:appendix:dataset}. The source code is publicly available \footnote{\url{https://stevetantan.github.io/HydraRAG/}}.


\vspace{2mm}
\myparagraph{Experiment baselines}
\label{exp:Baselines}
We compare \srag to four categories of baselines under an unsupervised setting with GPT-3.5-turbo as the LLM:
\begin{itemize}
    \item 
LLM-only methods without external knowledge, include standard prompting (IO), Chain-of-Thought prompting (CoT) \cite{wei2022cot}, and Self-Consistency prompting (SC) \cite{wang2022self} with six in-context examples;


    \item 
Vanilla RAG, covers text-based retrieval from entity documents and web-based retrieval from the top three web search results (title and snippets, same as the sample in Figure~\ref{fig:intro_demo});  
    \item 
KG-based RAG, includes Think-on-Graph (ToG) \cite{tog1.0sun2023think} and Paths-over-Graph (PoG) \cite{pogtan2025paths};  
    \item 
Hybrid RAG, consists of Chain-of-Knowledge (CoK) \cite{li2023chaincok} and Think-on-Graph-2.0 (ToG-2) \cite{tog2.0ma2024think}, which retrieve from both Wikipedia and Wikidata. 
\end{itemize}

\noindent For the statistics of existing SOTA, we directly refer to their results and those of other baselines reported in their paper for comparison.
Following prior studies \cite{tog1.0sun2023think, tog2.0ma2024think, pogtan2025paths, plan-on-graph, debated-on-graph}, we use exact match accuracy (Hits@1) as the evaluation metric. Recall and F1 scores are not used since knowledge sources are not limited to
document databases \cite{tog1.0sun2023think, tog2.0ma2024think, pogtan2025paths}. 


\vspace{2mm}
\myparagraph{Experiment implementation}
\label{exp:implementation}
All experiments use GPT-3.5-Turbo as the primary LLM. To demonstrate plug-and-play flexibility, we also run \srag with GPT-4-Trubo, Deepseek-v3, Llama-3.1-70B, and Llama-3.1-8B. Following ToG-2 and PoG, we set the temperature to 0.4 during evidence exploration (to increase diversity) and to 0 during path pruning and answer generation (to ensure reproducibility). 
We use SentenceBERT \cite{reimers-gurevych-2019-sentence} as the dense retrieval model (DRM).
The maximum generation length is 256 tokens. We fix $W_{\max}=3$, $D_{\max}=3$, $W_1=100$, and $W_2=20$ for evidence pruning. In evidence pruning, we use $\lambda_{\mathrm{sem}}=0.7$, $\alpha_{\mathrm{cross}}=0.7$, $(\rho_{\mathrm{KG}},\rho_{\mathrm{Wiki}},\rho_{\mathrm{Web}})=(1.0,0.8,0.7)$, and equal weights $\alpha_k=0.33$ for each feature $f_k$. 
\begin{table}[H]
\caption{
Statistics and license information for the datasets used in this paper.
$^\ast$ denotes that we utilize the sampled tests reported by existing
SOTA work for fairly comparing ~\cite {tog1.0sun2023think, pogtan2025paths, tog2.0ma2024think}.
}
\label{fig:appendix:dataset}
\setlength\tabcolsep{2 pt}
\resizebox{0.99\columnwidth}{!}{
\begin{tabular}{@{}ccccc@{}}
\toprule
Dataset & Answer Format & License & Test & Train \\ \midrule
ComplexWebQuestions (CWQ)$^\ast$ & Entity & Apache-2.0 & 1,000 & 27,734 \\
WebQSP & Entity/Number & MSR-LA & 1,639 & 3,098 \\
AdvHotpotQA & Entity/Number & CC BY-SA 4.0 & 308 & 2,312 \\
QALD10-en & Entity/Number & MIT & 333 & -- \\
Simple Questions$^\ast$ & Entity/Number & CC BY 3.0 & 1,000 & 14,894 \\
Zero-Shot RE & Entity/Number & CC BY-SA 4.0 & 3,724 & 147,909 \\
WebQuestions & Entity/Number & CC-BY 4.0  & 2,032 & 3,778 \\ \bottomrule
\end{tabular}}
\end{table}

\newpage
\onecolumn
\section{Case Study: Multi-Source Cross-Verified Interpretable Reasoning}
\label{castudy}
In this section, we present Tables~\ref{tab:casestudy-KG-web}-\ref{tab:casestudy-3-souces2} to illustrate how \srag combines evidence from the KG, Wikipedia, and the Web for cross-verified reasoning. Through case studies involving questions with multiple entities, and verification across KG-Wiki, KG-Web, and three-source combinations, we show how \srag generates transparent, faithful, and interpretable chains of facts to enhance LLM reasoning. Paths from different sources are color-coded. Showing \srag's effectiveness in multi-entity and multi-hop question answering by providing clear, understandable reasoning paths that support accurate answers.
\begin{center}
\begin{table*}[h]
\centering
\vspace{4mm}
\caption{Multi-source interpretable reasoning for “What is the nationality of the wrestler who sang on A Jingle with Jillian?”.Paths from KG, Wiki, and Web are color-coded: \textcolor{kgblue}{KG}, \textcolor{wikigreen}{Wiki}, \textcolor{weborange}{Web}.}
\label{tab:casestudy-KG-web}
\begin{tabularx}{\textwidth}{@{}p{3.3cm}X@{}}
\toprule
\textbf{Field} & \textbf{Content} \\ \midrule

\textbf{Question} & What is the nationality of the wrestler who sang on A Jingle with Jillian? \\
\textbf{Answer} & American \\
\textbf{Topic Entity} & \{A Jingle with Jillian\} \\

\toprule

\textbf{LLM Indicator} & "wrestler" – sang on – "A Jingle with Jillian" – has nationality – answer(nationality) \\

\textbf{Split Questions} & 
split\_question 1: What wrestler sang on "A Jingle with Jillian"? \\
& split\_question 2: What is the nationality of the performer of "A Jingle with Jillian"? \\

\toprule

\textbf{Source Evidence} &

\textbf{\textcolor{kgblue}{KG Path:}} \\
& \textcolor{kgblue}{\{A Jingle with Jillian\} $\xrightarrow{\text{music.album.artist}}$ \{Jillian Hall\} $\xrightarrow{\text{people.person.nationality}}$ \{\textbf{United States of America}\}} \\[2pt]

& \textbf{\textcolor{weborange}{Web Excerpt:}} \\
& \textit{"Jillian Hall was born on September 6, 1980, in Ashland, Kentucky, United States. She holds an American nationality."} \\

& \textbf{\textcolor{weborange}{Web (converted KG-path):}} \\
& \textcolor{weborange}{\{A Jingle with Jillian\} $\xrightarrow{\text{music.album.artist}}$ \{Jillian Hall\} $\xrightarrow{\text{birthplace}}$ \{Ashland, Kentucky, United States\} $\xrightarrow{\text{implies nationality}}$ \{\textbf{American}\}} \\

\toprule

\textbf{\srag Answer} & 
\textbf{answer:} \{\textcolor{kgblue}{\textbf{American}}\} 
\newline
\textbf{reason:} Both the knowledge graph and the web source independently support the answer \{\textbf{American}\}. The KG path directly connects "A Jingle with Jillian" to its artist, Jillian Hall, and further to her nationality as "United States of America". The web source confirms this by stating her birthplace is in the United States and she holds American nationality. After converting the web statement to a KG-style path, both sources cross-validate the answer, strengthening its correctness and trustworthiness.
\\ 
\bottomrule



\end{tabularx}
\end{table*}
\end{center}

\begin{table*}
\centering
\caption{Multi-source interpretable reasoning for “What movie was Logan Lerman in that was decorated by Barry Greaves?”. Paths from KG, Wiki, and Web are color-coded: \textcolor{kgblue}{KG}, \textcolor{wikigreen}{Wiki}, \textcolor{weborange}{Web}.}
\label{tab:casestudy-kg-wiki}
\begin{tabularx}{0.99\linewidth}{>{\raggedright\arraybackslash}p{3cm} X}
\toprule
\textbf{Field} & \textbf{Content} \\ 
\midrule
\textbf{Question} & What movie was Logan Lerman in that was decorated by Barry Greaves? \\
\textbf{Answer} & Fury \\
\textbf{Topic Entity} & \{Logan Lerman, Barry Greaves\} \\
\toprule
\textbf{LLM Indicator} & “Logan Lerman” – acted in – movie – decorated by – “Barry Greaves” \\
\textbf{Split Questions} & 
split\_question 1: What movie was Logan Lerman in? \newline
split\_question 2: Which person decorated the movie? \\
\toprule
\textbf{Source Evidence} & 
\textbf{\textcolor{kgblue}{KG Path:}} \newline
\textcolor{kgblue}{%
\{Logan Lerman\}
$\xrightarrow{\text{{film.actor.film}}}$
\{\textbf{Fury}\}
$\xleftarrow{\text{{film.film.film\_set\_decoration\_by}}}$
\{Barry Greaves\}} \newline

\textbf{\textcolor{wikigreen}{Wiki Excerpt:}} \newline
\textit{“Lerman next starred in David Ayer's World War II-set film, Fury; in the film, Lerman's character is an American soldier fighting Nazi forces.”} \newline

\textbf{\textcolor{wikigreen}{Wiki (converted KG-path):}} \newline
\textcolor{wikigreen}{%
\{Logan Lerman\}
$\xrightarrow{\text{{starred in}}}$
\{\textbf{Fury}\}} \newline
\textit{(The Wikipedia article confirms Logan Lerman's participation in \textbf{Fury}. The decoration by Barry Greaves is verified by the KG path.)} \\
\toprule

\textbf{\srag Answer} & 

\textbf{answer:} \{\textcolor{kgblue}{\textbf{Fury}}\} 
\newline
\textbf{reason:} Both the knowledge graph and Wikipedia independently identify \textbf{\{Fury\}} as the movie that Logan Lerman acted in. The KG path links Logan Lerman and Barry Greaves directly to the film, while Wikipedia supports Logan Lerman's involvement in the same work. Together these sources provide cross-validated evidence for the answer. \\

\bottomrule
\end{tabularx}
\end{table*}
\begin{table*}
\centering
\vspace{4mm}
\caption{Multi-source interpretable reasoning for “What member of the Republican Party fought in the Battle of Vicksburg?”. Paths from KG, Wiki, and Web are color-coded: \textcolor{kgblue}{KG}, \textcolor{wikigreen}{Wiki}, \textcolor{weborange}{Web}.}
\label{tab:3-soucr}
\begin{tabularx}{0.97\textwidth}{@{}p{3.3cm}X@{}}
\toprule
\textbf{Field} & \textbf{Content} \\ \midrule

\textbf{Question} & What member of the Republican Party fought in the Battle of Vicksburg? \\
\textbf{Answer} & Ulysses S. Grant \\
\textbf{Topic Entities} & \{Siege of Vicksburg, Republican Party\} \\

\toprule

\textbf{LLM Indicator} & "Siege of Vicksburg" – also known as – "Battle of Vicksburg" – fought by – answer(member) – member of – "Republican Party" \\

\textbf{Split Questions} & 
split\_question 1: What battle is also known as the "Siege of Vicksburg"? \\
& split\_question 2: What member of the "Republican Party" fought in the "Battle of Vicksburg"? \\

\toprule

\textbf{Source Evidence} &

\textbf{\textcolor{kgblue}{KG Path:}} \\
& \textcolor{kgblue}{
\{Siege of Vicksburg\} $\xrightarrow{\text{also known as}}$ \{Battle of Vicksburg\} $\xrightarrow{\text{fought by}}$ \{Ulysses S. Grant\} $\xrightarrow{\text{member of}}$ \{Republican Party\}
} \\[2pt]

&\textbf{\textcolor{wikigreen}{Wiki Excerpt:}} \\
& \textit{"The Siege of Vicksburg (May 18 – July 4, 1863) was the final major military action in the Vicksburg campaign of the American Civil War. In a series of maneuvers, Union Major General Ulysses S. Grant and his Army..."} \\
&\textbf{\textcolor{wikigreen}{Wiki (converted KG-path):}} \\
& \textcolor{wikigreen}{
\{Siege of Vicksburg\} $\xrightarrow{\text{also known as}}$ \{Battle of Vicksburg\} $\xrightarrow{\text{fought by}}$ \{Ulysses S. Grant\}
} \\[2pt]

&\textbf{\textcolor{weborange}{Web Excerpt:}} \\
& \textit{"The Battle of Vicksburg, or Siege of Vicksburg, was the final significant battle... Union Maj. Gen. Ulysses S. Grant and his Army of the Tennessee crossed the Mississippi River..."} \\
&\textbf{\textcolor{weborange}{Web (converted KG-path):}} \\
& \textcolor{weborange}{
\{Battle of Vicksburg\} $\xrightarrow{\text{fought by}}$ \{Ulysses S. Grant\}
} \\

\toprule

\textbf{\srag Answer} &
\textbf{answer:} \{\textcolor{kgblue}{\textbf{Ulysses S. Grant}}\} \\[2pt]
&\textbf{reason:} All three sources—KG, Wikipedia, and Web—support that Ulysses S. Grant fought in the Battle (Siege) of Vicksburg. The KG path further confirms his Republican Party membership. Wiki and Web sources confirm his role as a military leader in the battle, and after conversion to KG-path style, all sources consistently point to \textbf{\{Ulysses S. Grant\}} as the answer, demonstrating robust multi-source verification.\\
\bottomrule
\end{tabularx}
\end{table*}
\begin{table*}
\centering
\vspace{4mm}
\caption{Multi-source interpretable reasoning for “What team that has a mascot named Mariner Moose is in the American League West?”. Paths from KG, Wiki, and Web are color-coded: \textcolor{kgblue}{KG}, \textcolor{wikigreen}{Wiki}, \textcolor{weborange}{Web}.}
\label{tab:casestudy-3-souces2}
\begin{tabularx}{0.97\textwidth}{@{}p{3.3cm}X@{}}
\toprule
\textbf{Field} & \textbf{Content} \\ \midrule

\textbf{Question} & What team that has a mascot named Mariner Moose is in the American League West? \\
\textbf{Answer} & Seattle Mariners \\
\textbf{Topic Entities} & \{Mariner Moose, American League West\} \\

\toprule

\textbf{LLM Indicator} & "Mariner Moose" – mascot of – team – division – answer(team) – located in – "American League West" \\

\textbf{Split Questions} & 
split\_question 1: Which team has a mascot named "Mariner Moose"? \\
& split\_question 2: Which team is in the "American League West" division? \\

\toprule

\textbf{Source Evidence} &

\textbf{\textcolor{kgblue}{KG Path:}} \\
& \textcolor{kgblue}{
\{Mariner Moose\} $\xrightarrow{\text{sports.mascot.team}}$ \{Seattle Mariners\} $\xrightarrow{\text{baseball.baseball\_team.division}}$ \{American League West\}
} \\[2pt]

&\textbf{\textcolor{wikigreen}{Wiki Excerpt:}} \\
& \textit{"The Mariner Moose is the team mascot of the Seattle Mariners, a Major League Baseball team... The Seattle Mariners are an American professional baseball team based in Seattle. The Mariners compete in Major League Baseball (MLB) as a member club of the American League (AL) West Division."} \\
&\textbf{\textcolor{wikigreen}{Wiki (converted KG-path):}} \\
& \textcolor{wikigreen}{
\{Mariner Moose\} $\xrightarrow{\text{mascot of}}$ \{Seattle Mariners\} $\xrightarrow{\text{member of}}$ \{American League West\}
} \\[2pt]

&\textbf{\textcolor{weborange}{Web Excerpt:}} \\
& \textit{"Their mascot is the Mariner Moose. The Seattle Mariners are an American professional baseball team based in Seattle. The Mariners compete in Major League Baseball (MLB) as a member club of the American League (AL) West Division."} \\
&\textbf{\textcolor{weborange}{Web (converted KG-path):}} \\
& \textcolor{weborange}{
\{Mariner Moose\} $\xrightarrow{\text{team mascot of}}$ \{Seattle Mariners\} $\xrightarrow{\text{compete in}}$ \{American League West\}
} \\

\toprule

\textbf{\srag Answer} &
\textbf{answer:} \{\textcolor{kgblue}{\textbf{Seattle Mariners}}\} \\[2pt]
&\textbf{reason:} All three sources—KG, Wikipedia, and Web—consistently support that the \textbf{\{Seattle Mariners\}} have Mariner Moose as their mascot and are a team in the American League West division. The KG path provides a direct multi-hop link; the Wiki and Web evidence, after conversion to KG-path style, corroborate both the team and its division membership. This provides strong cross-source verification of the answer.\\
\bottomrule
\end{tabularx}
\end{table*}

\newpage
\section{Prompts}\label{appendix:prompt}
In this section, we detail the prompts required for our main experimental procedures.
\begin{center}
\begin{minipage}{0.7\columnwidth}
    \vspace{2mm}
        \centering
    \begin{tcolorbox}[title=Question Analysis Prompt Template]
    \label{prompt:Question Analysis}
        \small
    You will receive a multi-hop question, which is composed of several interconnected queries, along with a list of topic entities that serve as the main keywords for the question. Your task is to break the question into simpler parts, using each topic entity once and provide a Chain of Thought (CoT) that shows how the topic entities are related. Note: Each simpler question should explore how one topic entity connects to others or the answer. The goal is to systematically address each entity to derive the final answer.
\vspace{5pt}
    
    \texttt{In-Context Few-shot}
\vspace{5pt}
    
    Q: \{Query\} 
    
    Topic Entity: \{Topic Entity\} 
    
    A: 
    \end{tcolorbox}
    \vspace{2mm}

\end{minipage}
\end{center}
\begin{center}
\begin{minipage}{0.7\columnwidth}
    \vspace{2mm}
    \begin{tcolorbox}[title=Agentic Source Selector Prompt Template]~\label{prompt:AgenticSourceSelector}
        \small

You are a source selection agent. Your task is to decide the most appropriate knowledge source(s) to answer a user's question. You will be provided with up to three sources:
 Local Knowledge Graph (KG)
 Wiki Documents (Wiki) 
 Web Search (Web).
Follow these steps carefully:

I. Analyze the question thoroughly.

II. Prioritize KG if available:
If KG alone is sufficient, select KG.
If KG is incomplete, check if Wiki can fill the missing information. If so, combine KG with Wiki.
If neither KG nor Wiki suffices, include Web search.
    
III. If KG is unavailable:
    Choose between Wiki and Web based on recency and the likely completeness of the Wiki documents.

Clearly state your reasoning, and then indicate your decision using these actions:

action1 for KG

action2 for Wiki

action3 for Web

Noted, combinations allowed (e.g., [action1 + action2]).
\vspace{5pt}

\texttt{In-Context Few-shot}
\vspace{5pt}

Q: \{Query\} 

Provided sources: \{Provided sources\} 

Question analysis: \{Question analysis\}

A: 
    \end{tcolorbox}
    \vspace{1mm}
\end{minipage}
\end{center}

\vspace{1mm}
\begin{center}
\begin{minipage}{0.7\columnwidth}
    \vspace{2mm}
    \begin{tcolorbox}[title=From Paragraph to Knowledge Path Prompt Template]~\label{prompt:ParagraphtoKnowledge}
        \small
You will receive a multi-hop question, which consists of several interrelated queries, a list of subject entities as the main keywords of the question, three related questions and answers returned by Google search, and three online related search results from Google search.
Your task is to summarize these search results, find sentences that may be related to the answer, and organise them into a knowledge graph path for each paragraph.
Note that at least one path for each paragraph should contain the main topic entities.
Please answer the question directly in the format below: [{Brad Paisley} - enrolled at - {West Liberty State College} - transferred to - {Belmont University} - earned - {Bachelor's degree}]

\vspace{5pt}

\texttt{In-Context Few-shot}
\vspace{5pt}

Q: \{Query\} 

Topic Entity: \{Topic Entity\} 

Paragraph 1:\{Paragraph 1\}

Paragraph 2:\{Paragraph 2\}

Paragraph 3:\{Paragraph 3\}

A: 
    \end{tcolorbox}
    \vspace{1mm}
\end{minipage}
\end{center}

\vspace{1mm}

\begin{center}
\begin{minipage}{0.88\columnwidth}
    \noindent where \texttt{\{Skyline Indicator\}, \text{and} \{Split Question\}} are obtained in Section~\ref{sec:med:initial}. 
    \texttt{\{Existing Knowledge Paths\}} and \texttt{\{Candidate Paths\}} denote the retrieved reasoning paths, which are formatted as a series of structural sentences, \noindent where, $i$ and $j$ in $r_{1_i}, r_{1_i}$ represent the $i$-th, $j$-th relation from each relation edge in the clustered question subgraph.
\end{minipage}
\end{center}
\vspace{3mm}
\begin{minipage}{0.99\columnwidth}
    \centering
    $\{e_{0x},...,e_{0z}\}\to r_{1_i} \to \{e_{1x},...,e_{1z}\}\to \dots$ $\to r_{l_j} \to \{e_{lx},...,e_{lz}\}$ 
    \\ $\dots$\\
    $\{e_{0x},...,e_{0z}\}\to r_{1_i} \to \{e_{1x},...,e_{1z}\}\to \dots$ $\to r_{l_j} \to \{e_{lx},...,e_{lz}\}$,
\end{minipage}

\begin{center}
\begin{minipage}{0.7\columnwidth}
    \vspace{2mm}
    \begin{tcolorbox}[title=Refined Exploration Prompt Template]~\label{prmpt:prompts_Refined_Exploration}
        \small
Given a main question, an uncertain LLM-generated thinking Cot that considers all the entities, a few split questions that you can use and finally obtain the final answer, the associated accuracy retrieved knowledge paths from the Related\_path section, and the main topic entities Please predict the additional evidence that needs to be found to answer the current question,  then provide a suitable query for retrieving this potential evidence. and give the possible Chains of Thought that can lead to the predicted result in the same format below, by the given knowledge path and your own knowledge.

\vspace{5pt}

\texttt{In-Context Few-shot}
\vspace{5pt}

Q: \{Query\} 

Topic Entity: \{Topic Entity\} 

Skyline Indicator:\{Skyline Indicator\}

Split Question:\{Split Question\}

Existing Knowledge Paths:\{Existing Knowledge Paths\}

A: 
    \end{tcolorbox}
    \vspace{1mm}
\end{minipage}
\end{center}

\vspace{1mm}

\begin{center}
\begin{minipage}{0.7\columnwidth}
    \vspace{2mm}
    \begin{tcolorbox}[title=Predict Exploration Prompt Template]~\label{prompt:Predict_Exploration}
        \small
Using the main question, a possibly uncertain chain of thought generated by a language model, some related split questions, paths from the "Related\_paths" section, and main topic entities:
please first provide three predicted results,
and second offer three possible chains of thought that could lead to these results, using the provided knowledge paths and your own knowledge.
If any answers are unclear, suggest alternative answers to fill in the gaps in the chains of thought, following the same format as the provided examples.

\vspace{5pt}

\texttt{In-Context Few-shot}
\vspace{5pt}

Q: \{Query\} 

Topic Entity: \{Topic Entity\} 

Skyline Indicator:\{Skyline Indicator\}

Split Question:\{Split Question\}

Existing Knowledge Paths:\{Existing Knowledge Paths\}

A: 
    \end{tcolorbox}
    \vspace{2mm}
\end{minipage}
\end{center}


\begin{center}
    
\begin{minipage}{0.7\columnwidth}
    \centering
    \vspace{2mm}
    
    \begin{tcolorbox}[title=LLM-aware Paths Select Prompt Template]\label{LLMselect}
        \small
Given a main question, a LLM-generated thinking Cot that considers all the entities, a few split questions that you can use one by one and finally obtain the final answer, and the associated retrieved knowledge graph path, \{set of entities (with id start with "m.")\} -> \{set of relationships\} -> \{set of entities(with id start with "m.")\}, 
Please score and give me the top three lists from the candidate paths set that are highly likely to be the answer to the question.

\vspace{5pt}
    
\texttt{In-Context Few-shot}

\vspace{5pt}
Q: \{Query\} 

Skyline Indicator:\{Skyline Indicator\}

Split Question:\{Split Question\}

Candidate Paths:\{Candidate Paths\}

A: 
\end{tcolorbox}
\vspace{2mm}
\end{minipage}
\end{center}

\vspace{5pt}

\begin{center}
    
\begin{minipage}{0.7\columnwidth}
    \vspace{2mm}
    \begin{tcolorbox}[title=Path Refinement Prompt Template]\label{prompt:path_refine}
        \small
Given a main question, an uncertain LLM-generated thinking Cot that considers all the entities, a few split questions that you can use one by one and finally obtain the final answer, the associated accuracy retrieved knowledge paths from the Related paths section, and main topic entities.
Your task is to summarize the provided knowledge triple in the Related paths section and generate a chain of thoughts by the knowledge triple related to the main topic entities of the question, which will be used for generating the answer for the main question and splitting the question further.
You have to make sure you summarize correctly by using the provided knowledge triple, you can only use the entity with the id from the given path, and you can not skip steps.

\vspace{5pt}
    
\texttt{In-Context Few-shot}

\vspace{5pt}
Q: \{Query\} 

Skyline Indicator:\{Skyline Indicator\}

Split Question:\{Split Question\}

Related Paths:\{Related Paths\}

A: 
    \end{tcolorbox}
    \vspace{2mm}
\end{minipage}
\end{center}



\begin{center}
    
\begin{minipage}{0.7\columnwidth}
    \vspace{2mm}
    
    \begin{tcolorbox}[title= CoT Answering Evaluation Prompt Template]\label{prompt:corevaluation}
        \small
Given a main question, an uncertain LLM-generated thinking Cot that considers all the entities, a few split questions that you can use and finally obtain the final answer, and the associated retrieved knowledge graph path, \{set of entities (with id start with "m.")\} -> \{set of relationships\} -> \{set of entities(with id start with "m.")\}.
Your task is to determine if this knowledge graph path is sufficient to answer the given split question first then the main question. 
If it's sufficient, you need to respond \{Yes\} and provide the answer to the main question. If the answer is obtained from the given knowledge path, it should be the entity name from the path. Otherwise, you need to respond \{No\}, then explain the reason.

\vspace{5pt}
    
\texttt{In-Context Few-shot}

\vspace{5pt}
Q: \{Query\} 

Skyline Indicator:\{Skyline Indicator\}

Split Question:\{Split Question\}

Existing Knowledge Paths:\{Existing Knowledge Paths\}

A: 
    \end{tcolorbox}
    \vspace{2mm}
\end{minipage}
\end{center}

\begin{center}

\begin{minipage}{0.7\columnwidth}
    \vspace{2mm}
    \begin{tcolorbox}[title= CoT Answering Generation Prompt Template]\label{prompt:cot_gen}
        \small
Given a main question, an uncertain LLM-generated thinking Cot that considers all the entities, a few split questions that you can use one by one and finally obtain the final answer, and the associated retrieved knowledge graph path, \{set of entities (with id start with "m.")\} -> \{set of relationships\} -> \{set of entities(with id start with "m.")\}, 
Your task is to generate the answer based on the given knowledge graph path and your own knowledge.

\vspace{5pt}
    
\texttt{In-Context Few-shot}

\vspace{5pt}
Q: \{Query\} 

Skyline Indicator:\{Skyline Indicator\}

Split Question:\{Split Question\}

Related Paths:\{Related Paths\}

A: 
    \end{tcolorbox}
    \vspace{2mm}
\end{minipage}
\end{center}

\end{document}